\documentclass[times,twocolumn,final]{elsarticle}
\usepackage{medima}
\usepackage{framed,multirow}
\usepackage[table]{xcolor}
\usepackage{float}
\usepackage{amssymb}
\usepackage{latexsym}
\usepackage{booktabs}
\usepackage{dsfont}
\usepackage{algorithm}
\usepackage{algpseudocode}
\usepackage{url}
\usepackage{xcolor}
\usepackage{mathrsfs}
\usepackage{hyperref}
\usepackage{ragged2e}
\definecolor{red}{rgb}{0,0,0}
\definecolor{c1}{HTML}{6D8C00}
\definecolor{newcolor}{rgb}{.8,.349,.1}
\usepackage[normalem]{ulem}

\usepackage{amsmath}
\journal{Medical Image Analysis}
\begin{document}
\verso{Yanyan Wang \textit{et~al.}}
\begin{frontmatter}
\title{Leveraging Labelled Data Knowledge: A Cooperative Rectification Learning Network for Semi-supervised 3D Medical Image Segmentation}%
\centering

\author[1,3]{Yanyan \snm{Wang}}
\ead{wyanyan@stumail.neu.edu.cn}
\author[1]{Kechen \snm{Song}}
\ead{songkc@me.neu.edu.cn}
\author[2]{Yuyuan \snm{Liu}}
\ead{yuyuan.liu@adelaide.edu.au}
\author[1]{Shuai \snm{Ma}}
\ead{mashuai@stumail.neu.edu.cn}
\author[1]{Yunhui \snm{Yan}\corref{cor1}}
\ead{yanyh@mail.neu.edu.cn}
\author[3]{Gustavo \snm{Carneiro}\corref{cor1}}
\ead{g.carneiro@surrey.ac.uk}
\cortext[cor1]{Corresponding authors}

\address[1]{School of Mechanical Engineering and Automation, Northeastern University, China}
\address[2]{Australian Institute for Machine Learning, University of Adelaide, Australia}
\address[3]{Centre for Vision, Speech and Signal Processing (CVSSP), University of Surrey, UK}

\justify

\begin{abstract}
Semi-supervised 3D medical image segmentation aims to achieve accurate segmentation using few labelled data and numerous unlabelled data. 
The main challenge in the design of semi-supervised learning methods consists in the effective use of the unlabelled data for training.
A promising solution consists of ensuring consistent predictions across different views of the data, where the efficacy of this strategy depends on the accuracy of the pseudo-labels generated by the model for this consistency learning strategy.
In this paper, we introduce a new methodology to produce high-quality pseudo-labels for a consistency learning strategy to address semi-supervised 3D medical image segmentation.
The methodology has three important contributions.
The first contribution is the Cooperative Rectification Learning Network (CRLN) that 
learns multiple prototypes per class to be used as external knowledge priors to adaptively rectify pseudo-labels at the voxel level. 
The second contribution consists of the Dynamic Interaction Module (DIM) to facilitate pairwise and cross-class interactions between prototypes and multi-resolution image features, enabling the production of accurate voxel-level clues for pseudo-label rectification.
The third contribution is the Cooperative Positive Supervision (CPS), which optimises uncertain representations to align with unassertive representations of their class distributions, improving the model's accuracy in classifying uncertain regions.
Extensive experiments on three public 3D medical segmentation datasets demonstrate the effectiveness and superiority of our semi-supervised learning method. The code is available at \url{https://github.com/Yaan-Wang/CRLN.git}.
\end{abstract}


\end{frontmatter}

\section{Introduction}

3D medical image segmentation plays a crucial role in healthcare tasks by automatically identifying internal structures in medical volumes. While fully supervised segmentation has achieved outstanding performance~\citep{minaee2021image,wang2022medical}, it requires large amounts of voxel-wise annotations that are time-consuming and labour-intensive to acquire. In contrast, collecting large amounts of unlabelled samples is easier to obtain~\citep{li2020shape,lei2022semi}. Hence, semi-supervised medical image segmentation can relieve the annotation burden by utilising numerous unlabelled data, accompanied by a few labelled ones \citep{chen2022semi,wu2022mutual}. A key concern in this field is how to take advantage of unlabelled medical samples effectively \citep{luo2021semi,li2024contour}.

A classical approach is to employ a consistency learning strategy across different views of unlabelled data~\citep{wu2021semi}. 
An example of such strategy is the Mean Teacher (MT) model, which employs a "teacher-student" framework, where the student network generates predictions from the strongly augmented unlabelled data to converge towards the teacher's prediction of the weakly augmented version of the same unlabelled data~\citep{ADIGAV2024103011}. However, the pseudo-labels generated by the teacher may include incorrect predictions, which could adversely impact training, leading to the phenomenon known as confirmation bias~\citep{ZHANG2022369}. 
To address this problem, some methods employ multiple auxiliary teacher models that are initialized and updated differently, thereby generating diverse pseudo-labels for the distinct views~\citep{zhao2023alternate}. 
The combination of such diverse pseudo-labels tends to yield more robust final labels, implicitly mitigating the confirmation bias. 
A potential drawback of employing multiple auxiliary teacher models is the challenge they may face in consistently enhancing the robustness of pseudo-labels across diverse views. This inconsistency stems from the varying predictions of different views, which may not consistently align with the true ground truth. 

\begin{figure}[!t]
\centering
\includegraphics[width=3.5in]{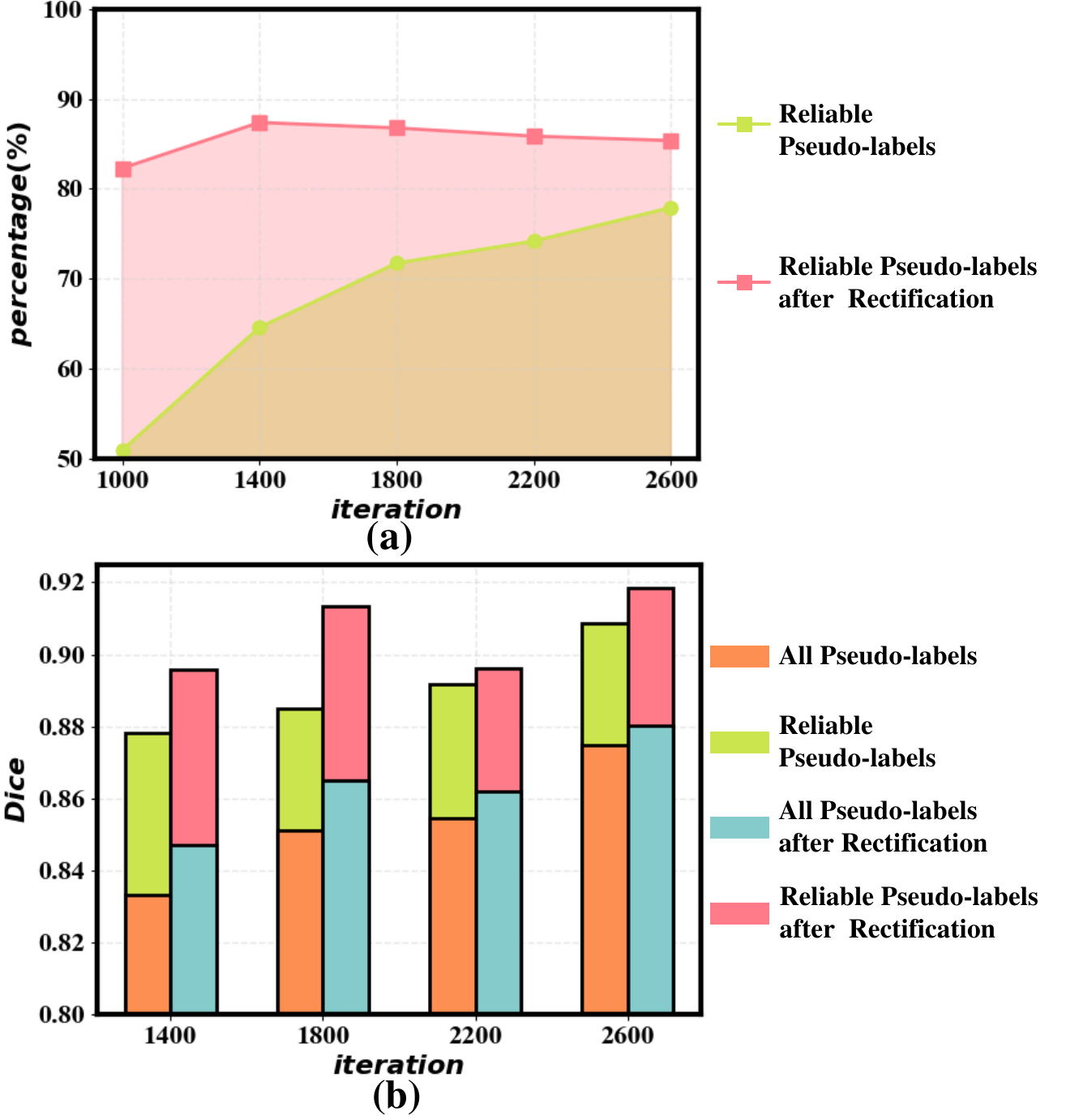}
\caption{Quality of pseudo-labels as a function of training iteration under the 10\% partition semi-supervised learning protocol on the LA dataset. (a) Proportion of reliable predictions within all pseudo-labels before (green curve) and after (red curve) our proposed rectification. (b) Segmentation accuracy (Dice) results of all pseudo-labels (orange), reliable pseudo labels (green), all rectified pseudo labels (blue), and reliable rectified pseudo labels (red). During training, especially in the early stages, only a small percentage of the predictions of the pseudo-labels are reliable, which produces relatively inaccurate segmentation. After applying our proposed rectification, the percentage of reliable pseudo-labels and their respective segmentation accuracy are significantly improved (Sec.~\ref{sec:Ablation_Study} has more details).}
\label{fig:quality_pseudo_label}
\end{figure}

Another widely adopted strategy to alleviate confirmation bias involves filtering out the noisy regions based on prediction uncertainty~\citep{luo2021efficient,yu2019uncertainty}. As illustrated in Fig.~\ref{fig:quality_pseudo_label}(a), the percentage of reliable predictions (i.e., predictions with low uncertainty measures) from pseudo-labels is not high during training, especially in the early stages. 
This means that a large number of unlabelled training data containing unreliable predictions are discarded, so they are not used for training. 
An additional problem is that voxels classified as reliable may in fact contain wrong predictions, as shown in Fig.~\ref{fig:quality_pseudo_label}(b) (see results for 'Reliable Pseudo-labels'), which may result in confirmation bias. 

\begin{figure}[!t]
\centering
\includegraphics[width=3.5in]{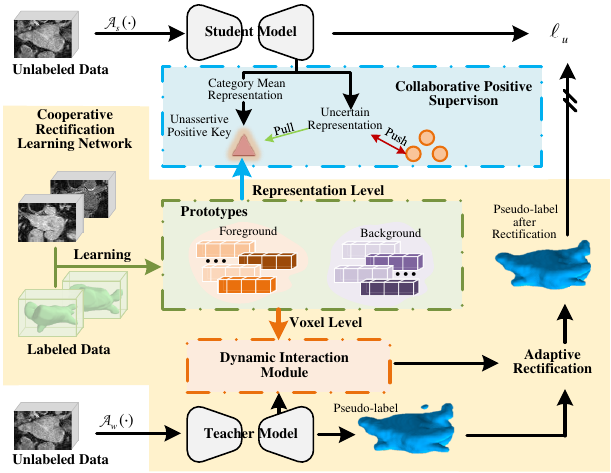}
\caption{The main contributions of this paper are the Cooperative Rectification Learning Network (CRLN, yellow box), Dynamic Interaction Module (DIM, pink box), and the Collaborative Positive Supervision (CPS, blue box). CRLN learns multiple class-wise prototypes that work as knowledge priors for the rectification of pseudo-labels.
DIM aims to acquire holistic relationships across multiple class prototypes and
unlabelled data to rectify pseudo labels.
CPS trains uncertain representations to get closer to their unassertive positive learning key, enabling the model to better discriminate such uncertain regions.
\label{fig:cooperative_rectification}}
\end{figure}

In this paper, we introduce a new methodology to improve the prediction accuracy of pseudo-labels in consistency-based semi-supervised learning approaches, thereby addressing the limitations mentioned above.
The methodology, shown in Fig.~\ref{fig:cooperative_rectification}, has three important contributions. 
The first contribution is the Cooperative Rectification Learning Network (CRLN) that leverages the prior information present in the learned set of prototypes for each segmentation class. These prototypes are used to adaptively rectify pseudo-labels at the voxel level.
The second contribution is the Dynamic Interaction Module (DIM) that provides accurate voxel-level clues for pseudo-label rectification by exploring pairwise and cross-class interactions between prototypes and multi-resolution image features. 
CRNL and DIM are designed to improve the quality of pseudo-labels, allowing more unlabelled data to be used during training, as shown by the methods after rectification in  Fig.~\ref{fig:quality_pseudo_label}.
Furthermore, semi-supervised segmentation models usually exhibit low classification confidence in discriminating samples located at uncertain regions of the feature space, such as at \textcolor{red}{class boundary regions.} To enhance the discrimination of uncertain regions, a Collaborative Positive Supervision (CPS) mechanism is proposed as the third contribution of this paper.
CPS is a contrastive learning method, where the main innovation is in the definition of the positive representations based on an unassertive class representation that combines the learned prototypes and the mean class representation, as opposed to the more common positive representation based on the mean value of the class representations~\citep{liu2021bootstrapping}.

To summarise, our main contributions are: 
\begin{enumerate}
    \item The new CRLN method that learns multiple prototypes per class to be explored as external knowledge priors for the rectification of pseudo-labels. 
    \item The novel DIM module devised to capture holistic relationships across multiple class prototypes and unlabelled data in a pairwise and cross-class manner, providing critical clues to rectify pseudo labels to improve their accuracy. 
    \item The innovative CPS mechanism 
    to encourage the uncertain representations to move closer to their positive representations, defined by an unassertive class representation that combines the learned prototypes and the mean class representation, which gives the model the ability to distinguish the uncertain regions. 
\end{enumerate}
We show that our semi-supervised learning method produces the best result in the field on three public 3D medical segmentation datasets, namely:  the Left Atrium (LA) \citep{xiong2021global}, Pancreas-CT \citep{clark2013cancer}, and Brain Tumour Segmentation 2019 (BraTS19) \citep{menze2014multimodal}.
The remainder of the paper is organised as follows: Section ~\ref{sec:Related Work} provides an overview of medical image segmentation and semi-supervised medical image segmentation. Then, the details of the proposed method are presented in Section ~\ref{sec:Method}. Section ~\ref{sec:Experiments} illustrates experimental results and relevant analysis. Finally, Section ~\ref{sec:Conclusion} concludes this paper.

\section{Related Work}\label{sec:Related Work}

\subsection{Medical Image Segmentation}

Medical image segmentation aims to assign a closed-set class label to each pixel. UNet \citep{ronneberger2015u} has gradually become the preferred model in the medical segmentation field since it was first proposed in 2015. Such success is mainly attributed to its unique structural design, especially the encoder-decoder structure and the skip connection mechanism, which endows the model with strong detail recovery. Subsequently, it is enhanced by exploring dense skip connections~\citep{guan2019fully,zhou2018unet++}, multi-scale receptive fields~\citep{xiao2018weighted} and global information~\citep{chen2021transunet}. 
In addition to 2D medical scenes, UNet has been extended to 3D medical scenes, such as MRI and CT, by replacing 2D convolutions with 3D convolutions~\citep{cciccek20163d}. Another similar approach is VNet \citep{milletari2016v}, which also employs the encoder-decoder structure. The difference with 3D-UNet is that VNet utilises a convolutional layer instead of the pooling layer for downsampling, thereby mitigating information loss. 

More recent studies~\citep{10093768,horst2024cellvit} have explored methods based on vision transformer (ViT)~\citep{dosovitskiy2020image} for the medical segmentation task, leveraging their capability for long-range modelling. Other recent approaches~\citep{chowdary2023diffusion,wu2024medsegdiff} can enhance the segmentation quality by introducing the Diffusion Probabilistic Model (DPM)~\citep{ho2020denoising}. 
Although the fully supervised segmentation techniques described above exhibit solid performance, they require a large number of voxel-based annotations that are difficult and expensive to obtain in real medical scenarios. 
One way to mitigate the need for such annotations is based on the development of semi-supervised learning methods that require much smaller sets of annotated data and large sets of un-annotated data. We review semi-supervised learning methods below.


\subsection{Semi-supervised Medical Image Segmentation}

Significant advancements have been made in semi-supervised medical image segmentation~\citep{miao2023caussl, wang2023mcf}. Current models primarily adopt consistency regularisation strategies to leverage unlabelled information~\citep{hang2020local, wang2020double}. The underlying idea behind these strategies is that the model's predictions for unlabelled samples should remain consistent under various perturbations~\citep{bai2023bidirectional, gao2023correlation}. The Mean Teacher framework, which explores weak and strong augmentation strategies, is quite effective in the implementation of this idea. Based on this framework, various weak and strong augmentation techniques have been proposed to generate prediction disagreements. \citet{li2020transformation} apply different data augmentation techniques, such as Gaussian noise and contrast variation, on the input data. \citet{liu2022translation} adjust the spatial context of the input samples to enrich their diversity. Also, \citet{xu2022learning} and \citet{zheng2022double} focus on inducing prediction inconsistencies by adding perturbations at the feature level.

Despite the promising performance of well-designed data augmentation techniques, the pseudo-labels generated by teacher networks still contain a fair amount of noise, which hinders the model's segmentation capability. Recent works argue that incorporating additional supervised information would help mitigate this problem. One of the representative efforts is the multi-teacher embedding approach~\citep{liu2022perturbed, zhao2023alternate}. The core idea of this kind of approach lies in the generation of pseudo-labels from different perspectives. To ensure diversity, the teacher models typically employ different initialization parameters and update mechanisms. However, not all perspectives necessarily improve the accuracy of pseudo-labels, and sometimes conflicting labels may emerge. It is thus difficult to utilise different perspectives from unlabelled information to improve segmentation performance in complex situations. 

\begin{figure*}[!t]
\centering
\includegraphics[width=6.5in]{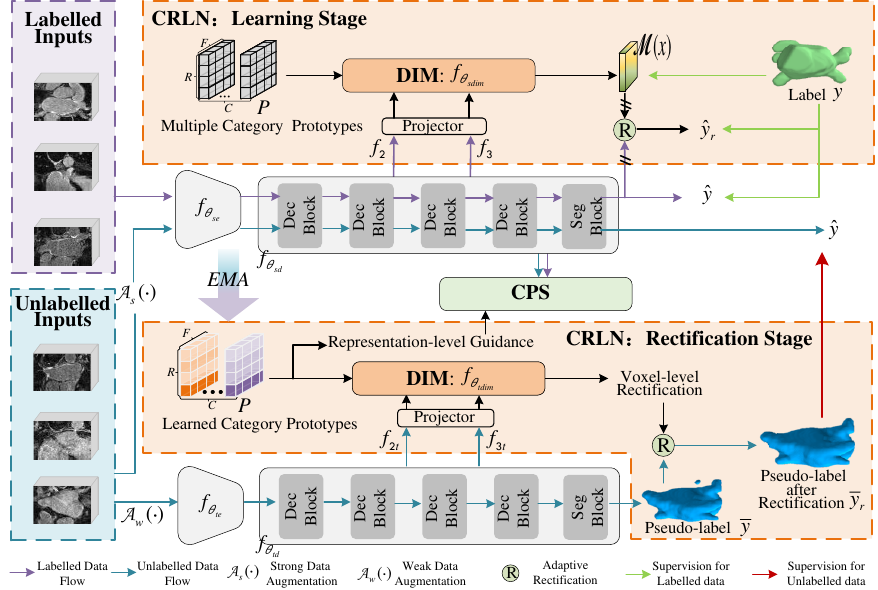}
\caption{The architecture of the proposed model. Based on the teacher-student structure, the Cooperative Rectification Learning Network (CRLN) consists of two stages. In the learning stage, multiple category prototypes are built and initialised. Subsequently, the Dynamic Interaction Module (DIM) implements pairwise interactions, as well as spatial-aware and cross-class aggregation between prototypes and the semantics of the labelled data to obtain the holistic relationship map $\mathscr{M}(x)$ which adaptively improves the segmentation quality of $\hat{y}$ with~\eqref{eq:rectification_student}. By minimising the deviation between predictions and labels, the proposed CRLN effectively learns valuable category prototypes and understands how to use them for voxel-level correction. In the rectification stage, the learned category prototypes serve as prior knowledge to rectify the pseudo-labels $\bar{y}$. After rectification, the higher-quality pseudo-labels $\bar{y}_{r}$ are used as supervision signals. Moreover, the Collaborative Positive Supervision (CPS) mechanism constructs unassertive centres by integrating learned category prototypes and category mean representations, allowing for better contrastive learning ($\ell_{cp}(\cdot)$ in ~\eqref{eq:cp} of representations with lower predictive confidence in the student network. This empowers the model to distinguish uncertain regions.
\label{fig:network_structure}}
\end{figure*}

Another technique to suppress the negative effects of noise in pseudo-labels is to filter out or reduce the weight of samples classified as uncertain during training~\citep{wang2021tripled,xia2020uncertainty}. UA-MT estimates the uncertainty of the teacher's prediction with the classification entropy and uses only reliable (i.e., low-entropy) predictions to supervise the student network~\citep{yu2019uncertainty}. \citet{luo2021efficient} propose to use multi-scale prediction discrepancy as a measure of uncertainty and then treat the uncertainty score as a pixel-level coefficient to reduce the loss contribution from the uncertain regions. \citet{luo2021semi} perform uncertainty estimation via subjective logic. Furthermore, \citet{su2024mutual} compute the reliability of the pseudo-labels based on the intra-class consistency. 
The computation of these uncertainty maps focuses on pixel-level local information and overlooks the benefits of the global view. Therefore, \citet{ADIGAV2024103011} employ a denoising autoencoder to reconstruct the predictions of the teacher network and then implement uncertainty estimation by calculating the difference between the teacher model's prediction and its reconstruction. \textcolor{red}{Recently, \citet{10273222} propose to adaptively weight the pseudo supervision loss in a voxel-wise manner based on the uncertainty of model predictions and promote feature consistency across differently augmented samples using a contrastive loss. These methods, which filter out or reduce the loss weight of voxels classified as uncertain, can reduce the negative impact of noise in pseudo-labels. However, voxels classified as reliable may, in fact, contain incorrect predictions, potentially harming model performance. Moreover, many regions in the pseudo-labels have low predictive confidence during training, especially in the early training stages, leading to the under-utilization of both these labels and their corresponding raw unlabelled data. Hence, we propose to leverage the labelled knowledge to explicitly rectify the low-quality predictions present in the pseudo-labels.}
\begin{figure}[!t]
\centering
\includegraphics[width=3.5in]{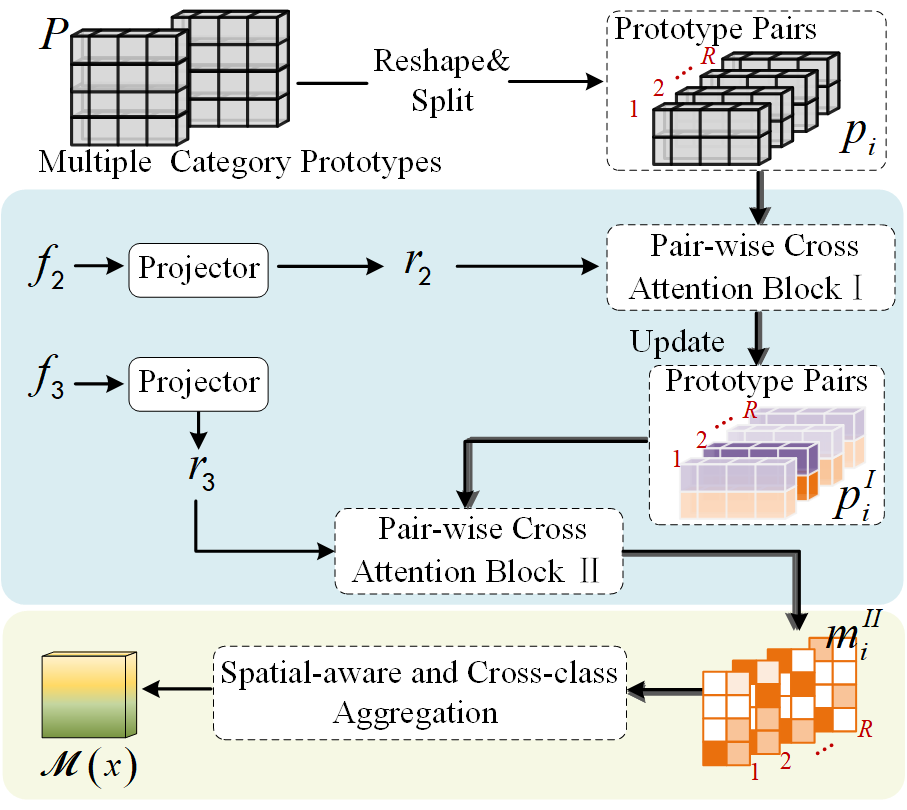}
\caption{\textcolor{red}{The architecture of the Dynamic Interaction Module (DIM).}}
\label{fig:dim}
\end{figure}

\section{Method}\label{sec:Method}

The semi-supervised 3D medical image segmentation task aims to achieve precise organ segmentation utilising a limited number of labelled samples and a large number of unlabelled samples. Let $\mathcal{L}=\left\{x_i,y_i\right\}_{i=1}^L$ represents the labelled dataset, where $x \in \mathcal{X} \subset \mathbb{R}^{H \times W\times D}$ is the input volume, and $y \in \mathcal{Y} = \{0,1 \}^{C \times H \times W \times D}$ is the voxel-wise one-hot label within the $C$ classes. Additionally, the unlabelled dataset is denoted by $\mathcal{U}=\left\{x_i\right\}_{i=1}^U$, where $U>>L$. As shown in Fig.~\ref{fig:network_structure}, our method is built upon the teacher-student paradigm, so we begin with a brief review of its workflow in Sec.~\ref{sec:Preliminaries}. Next, the Cooperative Rectification Learning Network (CRLN) and Dynamic Interaction Module (DIM) are introduced in Sec.~\ref{sec:pdr}, and the Collaborative Positive Supervision (CPS) is explained in Sec.~\ref{sec:cps}. Finally, Sec.~\ref{sec:Holistic_Training_Objective} summarises the overall training objective.\par

\subsection{Preliminaries}\label{sec:Preliminaries}
The teacher-student paradigm employs a weak-to-strong consistency regularisation to train the segmentation model, allowing for simultaneous utilisation of labelled and unlabelled samples. More concretely, each unlabelled volume $x_i \in \mathcal{U}$ undergoes a weak augmentation process, denoted by $\mathscr{A}_w:\mathcal{X} \to \mathcal{X}$ (e.g., random cropping and random flipping), as well as strong augmentation $\mathscr{A}_s:\mathcal{X} \to \mathcal{X}$ (e.g., random cropping, random flipping, random noise and CutMix~\citep{yun2019cutmix}). Subsequently, the strongly augmented volume is fed into the student model, whereas the weakly augmented volume is fed to the teacher model to generate the pseudo-label. So the model can be optimised by the following objective:
\begin{equation}
\begin{split}
\ell(\mathcal{L},\mathcal{U},\theta)= & \frac{1}{L} \sum_{(x,y)\in\mathcal{L}} \ell_s \left(f_{\theta_{sd}}\left(f_{\theta_{se}}(x) \right),y\right) + \\ & \frac{1}{U} \sum_{(x)\in\mathcal{U}} \ell_u\left (f_{\theta_{sd}}\left (f_{\theta_{se}}(\mathscr{A}_s(x))\right),\textcolor{red}{\mathscr{A}_{g}} \left( f_{\theta_{td}}\left(f_{\theta_{te}}(\mathscr{A}_w(x))\right)\right)\right),
\end{split}
\label{eq:training_loss}
\end{equation}
where $\theta=\{ \theta_{sd}, \theta_{se}, \theta_{td}, \theta_{te} \}$ represents the student and teacher model parameters, 
$f_{\theta_{sd}}(f_{\theta_{se}}(x))$ denotes the student predictor (with $f_{\theta_{sd}}:\mathcal{F} \to [0,1]^{C \times H \times W \times D}$ denoting the student decoder,  $f_{\theta_{se}}:\mathcal{X} \to \mathcal{F}$ representing the student encoder, and $\mathcal{F} \in \mathbb{R}^{F}$),  $f_{\theta_{td}}(f_{\theta_{te}}(x))$ represents the teacher predictor (similarly, $f_{\theta_{td}}:\mathcal{F} \to [0,1]^{C \times H \times W \times D}$ is the teacher decoder and $f_{\theta_{te}}:\mathcal{X} \to \mathcal{F}$ is the teacher encoder), \textcolor{red}{$\mathscr{A}_{g}:[0,1]^{C \times H \times W \times D} \to [0,1]^{C \times H \times W \times D}$ represents the geometric transformation used to align the teacher's and student's predictions}, $\ell_s(\cdot)$ is the supervised learning term (e.g., Dice loss and cross-entropy loss), and the unsupervised learning loss is defined by
\begin{equation}
\label{eq:filter_teacher}
\ell_u(\hat{y},\bar{y}) = \frac{1}{|\Omega|}\sum_{\omega \in \Omega} \mathds{1}\left(\max_{\hat{c} \in \{1,...,C\} } \left (\bar{y}(\omega,\hat{c})\right ) \geq \tau \right) \times \ell_{nll}\left(\hat{y}(\omega,:),\bar{y}(\omega,:)\right),
\end{equation}
with $\Omega$ denoting the image lattice, $\hat{y} = f_{\theta_{sd}}(f_{\theta_{se}}(\mathscr{A}_s(x)))$,  $\bar{y}=f_{\theta_{td}}(f_{\theta_{te}}(\mathscr{A}_w(x)))$, $\ell_{nll}(\cdot)$ being the negative log-likelihood loss, and $\mathds{1}(\max_{\hat{c} \in \{1,...,C\} }(\bar{y}(\omega,\hat{c})) \geq \tau)$ representing the indicator function that filters out uncertain predictions where the maximum probability is smaller than $\tau \in [0,1]$.

Both the student and teacher models share the same architecture, with the weights of the teacher model updated by the Exponential Moving Average (EMA) of the student model weights.


\subsection{Pseudo-label Rectification}\label{sec:pdr}

The filtering mechanism described in~\eqref{eq:filter_teacher} effectively addresses the adverse effects of uncertainty in the pseudo-labels. However, an excessive filtration process runs the risk of underutilising the valuable information present in the unlabelled training set. 
To address this issue, we propose a method to rectify the uncertain predictions of the pseudo-labels by leveraging the knowledge present in the labelled data. Such rectification will allow the use of a larger number of unlabelled training samples, thereby improving the generalisation of the model.



Inspired by recent progress shown in the development of transformer-based architectures\citep{li2023transformer} and few-shot learning \citep{liu2022dynamic}, we propose the Cooperative Rectification Learning Network (CRLN) to learn discriminative class prototypes from labelled data that are then used for the rectification of predicted pseudo-labels. 
As depicted in Fig.~\ref{fig:network_structure}, CRLN comprises a prototype learning stage and a rectification stage. 
Moreover, to enhance the learning of prototypes and provide accurate voxel-level cues for correction, the Dynamic Interaction Module (DIM) is devised to take pairwise interactions, as well as spatial-aware and cross-class aggregation between prototypes and labelled data (in the learning stage) or the unlabelled data (in the rectification stage).
 

\subsubsection{Learning Stage} 
\textcolor{red}{\textbf{Learning Multiple Class Prototypes.}} Class prototypes are denoted by vectors in the feature space $\mathcal{F}$ of the decoder, offering a compressed representation of the class distribution. Since medical scenarios are complex and variable, it is difficult for a single prototype to cover the richness of the class representation. Therefore, we propose a method that represents a class with multiple prototypes. The prototypes are denoted by $P \in \mathbb{R}^{C \times R \times F}$, where $C$ is the number of classes, $R$ represents the number of prototypes for each class and $F$ is the dimensionality of the feature space $\mathcal{F}$ of the decoder.

These multiple prototypes per class are estimated during the learning stage of the CRLN, which also depends on the DIM module. \textcolor{red}{To capture the most relevant features of each class and maintain diversity among multiple prototypes from a global perspective, we propose first grouping multiple prototypes into $R$ matrices of size $C \times F$, with each matrix containing one prototype per class, and then gradually interacting these prototype matrices with the features through cross-attention. As illustrated in Fig.~\ref{fig:dim}, the prototypes $P$ are divided into $R$ sets $\left\{p_{i}\right\}_{i=1}^{R}$, with each $p_{i} \in \mathbb{R}^{C \times F}$ containing $C$ prototypes.}
The feature $f_{2}\in \mathbb{R}^{H/4 \times W/4 \times D/4 \times F}$ extracted from the second intermediate layer of the student decoder is processed by a $1 \times 1 \times 1 $ convolution layer to produce $r_{2} = \mathsf{Conv}_{1\times 1\times 1}(f_{2})$.  
These transferred contexts $r_{2}$ then interact with each prototype matrix in $\left\{p_{i}\right\}_{i=1}^{R}$ by the pair-wise cross attention block \uppercase\expandafter{\romannumeral1},  formulated as
\begin{equation}
\label{eq:update_prototypes1}
m^{\uppercase\expandafter{\romannumeral1}}_{i}= \frac{\phi^{\uppercase\expandafter{\romannumeral1}}_q(p_{i})\phi^{\uppercase\expandafter{\romannumeral1}}_k({r_{2}})^{\textcolor{red}{\top}}}{\sqrt{F}}, i \in \{1,...,R\}, \text{ followed by }
\end{equation}
\begin{equation}
\label{eq:update_prototypes2}
p^{\uppercase\expandafter{\romannumeral1}}_{i}= \mathsf{softmax}(m^{\uppercase\expandafter{\romannumeral1}}_{i}) \phi^{\uppercase\expandafter{\romannumeral1}}_v({r_{2}}), i \in \{1,...,R\},
\end{equation}
where $\phi^{\uppercase\expandafter{\romannumeral1}}_q(\cdot)$, $\phi^{\uppercase\expandafter{\romannumeral1}}_k(\cdot)$ and $\phi^{\uppercase\expandafter{\romannumeral1}}_v(\cdot)$ are the linear projectors of block $\uppercase\expandafter{\romannumeral1}$, {\textcolor{red}{$^\top$}} denotes the matrix transpose operation, and $m^{\uppercase\expandafter{\romannumeral1}}_{i} \in \mathbb{R}^{C \times H/4 \times W/4 \times D/4}$.

\textcolor{red}{Subsequently, the updated prototype matrices $\left\{p^{\uppercase\expandafter{\romannumeral1}}_{i}\right\}_{i=1}^{R}$ interact with the features $f_{3}\in \mathbb{R}^{H/2 \times W/2 \times D/2 \times F_3}$ that are extracted from the third intermediate layer of the student decoder, where $F_3 < F$.}
Such interaction is processed by the pair-wise cross attention block \uppercase\expandafter{\romannumeral2}, in the same way as in~\eqref{eq:update_prototypes1}, to generate the proximity matrices defined as 
\begin{equation}
m^{\uppercase\expandafter{\romannumeral2}}_{i} =\frac{\phi^{\uppercase\expandafter{\romannumeral2}}_q(p^{\uppercase\expandafter{\romannumeral1}}_{i})\phi^{\uppercase\expandafter{\romannumeral2}}_k({r_{3}})^{\textcolor{red}{\top}}}{\sqrt{F_3}}, 
\label{eq:proximity_matrices}
\end{equation} 
where $i \in \{1,...,R\}$, $r_{3}= \mathsf{Conv}_{1\times 1\times 1}(f_{3})$, and $m^{\uppercase\expandafter{\romannumeral2}}_{i} \in \mathbb{R}^{ C \times H/2 \times W/2 \times D/2}$. \textcolor{red}{By interacting with features from various decoder layers through the repeating pair-wise cross attention operations, the class prototypes can progressively absorb both texture and semantic information, thus continuously refining their representations.}

The computed proximity matrices are used to produce the holistic relationship map that is taken as clues to rectify the pseudo-label predictions. A simple solution would be to produce these clues by summing the proximity matrices, but that would not allow us to account for dependencies between classes.
Specifically, we first incorporate spatial consistency into our approach by re-evaluating the relationships between feature points and each prototype group based on the local context by applying a $3 \times 3\times3$ convolution layer 
to $M^{\uppercase\expandafter{\romannumeral2}} \in \mathbb{R}^{C \times R \times N}$ (with $N=H/2\times W/2 \times D/2$), 
which is a tensor containing the proximity matrices $\left\{m^{\uppercase\expandafter{\romannumeral2}}_{i}\right\}_{i=1}^{R}$, with $m^{\uppercase\expandafter{\romannumeral2}}_{i} \in \mathbb{R}^{C \times N}$ computed from~\eqref{eq:proximity_matrices}.

\textcolor{red}{Then, we synthesise the relationships across the $R$ prototypes within each class using a $1 \times 1 \times 1 $ convolution layer that is more flexible than just summing tensor over the $R$ dimensions to produce the holistic relationship map.}
It is important to note that the convolutional parameters are shared between different classes to enable information interaction across classes. The whole process is formulated as
\begin{equation}
\label{eq:spatial_aware_cross_class_aggregation}
\mathscr{M}(x)=\mathsf{Upsample}\left(\mathsf{Conv}_{1 \times 1 \times 1}\left (\mathsf{Conv}_{3 \times 3 \times 3}(M^{\uppercase\expandafter{\romannumeral2}})\right)\right),
\end{equation}
where 
$\mathscr{M}(x) \in \mathbb{R}^{  C \times H \times W \times D  }$ represents the holistic relationship map. 
We denote all learnable parameters in~\eqref{eq:update_prototypes1},~\eqref{eq:update_prototypes2},~\eqref{eq:proximity_matrices},~\eqref{eq:spatial_aware_cross_class_aggregation} as $\theta_{dim}$, where the whole DIM is denoted by
\begin{equation}
f_{\theta_{dim}}:\mathcal{X} \to \mathbb{R}^{C \times H \times W \times D }.
\label{eq:dim}   
\end{equation}

\textcolor{red}{\textbf{Learning to Rectify.}} The holistic relationship map from the DIM, defined in~\eqref{eq:spatial_aware_cross_class_aggregation}, serves to dynamically rectify each voxel from the original prediction $\hat{y}$ with 
\begin{equation}
\label{eq:rectification_student}
\hat{y}_{r}= \hat{y} + (1-\mu)\times \mathscr{M}(x),
\end{equation}
where $\mu \in [0,1]$, as shown in Fig.~\ref{fig:network_structure}, is a learnable parameter utilised to regulate the extent of the correction and $\hat{y} = f_{\theta_{sd}}(f_{\theta_{se}}(\mathscr{A}_s(x)))$, with $x$ belonging to the labelled set $\mathcal{L}$. 


\subsubsection{Rectification Stage} 
\label{sec:rectification_stage}

After $S$ learning iterations, we rectify the pseudo-labels of the unlabelled samples in $\mathcal{U}$
with the learned prototypes $P \in \mathbb{R}^{C \times R \times F}$. 
The features at different scales of unlabelled samples follow the same progressive method in the learning stage to interact with the multiple prototypes from labelled samples to generate the holistic relationship map $\mathscr{M}(x)$. \textcolor{red}{Since this progressive interaction integrates the advantages of features from multiple layers, it provides accurate voxel-level cues for correcting the pseudo-labels.}
Then, the original pseudo-labels $\bar{y}=f_{\theta_{td}}(f_{\theta_{te}}(\mathscr{A}_w(x)))$, with $x \in \mathcal{U}$, can be rectified by~\eqref{eq:rectification_student}, thus generating higher-quality pseudo-labels $\bar{y}_{r} = \bar{y} + (1-\mu)\times \mathscr{M}(x)$.

\begin{figure}[!t]
\centering
\includegraphics[width=3.5in]{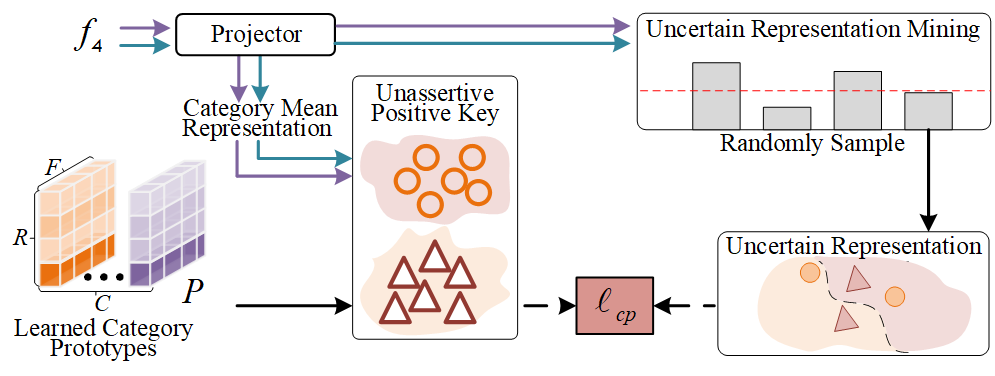}
\caption{\textcolor{red}{The architecture of the Collaborative Positive Supervision (CPS) mechanism.}}
\label{fig:cps}
\end{figure}

\subsection{Collaborative Positive Supervision}\label{sec:cps}

As discussed in Sec.~\ref{sec:pdr}, the CRLN leverages labelled data as priors to rectify pseudo-labels in voxel space, thus providing more reliable supervision for model training. Nonetheless, the model may still exhibit low confidence in segmenting challenging regions, such as \textcolor{red}{edge regions} and low-contrast regions. Therefore, we aim to mitigate this problem by enhancing the discriminative characteristics of the representations extracted from such challenging regions with contrastive learning. 
However, considering the practicality of the model's prediction of uncertain regions, we propose a more moderate contrastive learning strategy based on the InfoNCE loss \citep{he2020momentum}, namely the Collaborative Positive Supervision (CPS) mechanism. 

As shown in Fig.~\ref{fig:cps}, the proposed CPS mechanism is applied to the representations, extracted from the fourth layer of the student decoder, denoted by $f_{4} \in \mathbb{R}^{H\times W\times D \times F_4}$, where $F_4 < F_3$. ~\textcolor{red}{We first construct the anchor, negative and positive sets from $f_{4}$ for the InfoNCE loss, as explained below.}

\textcolor{red}{\textbf{Anchor Set.} For each class $c$, we randomly sample anchor points from uncertain regions of the feature space to form the class-specific set:}
\begin{equation}
\label{eq:anchor_set}
\textcolor{red}{
\mathcal{R}_c = \left\{ r(\omega,:) \mid \omega \in \Omega_c, \, r = \mathsf{Conv}_{1 \times 1 \times 1} \left(\mathsf{Conv}_{3 \times 3 \times 3} (f_{4})\right) \right\}},
\end{equation}
\textcolor{red}{where $\Omega_c$ denotes a subset of the lattice of size $H \times W \times D$ in the feature space. For labelled data in $\mathcal{L}$, this lattice subset is defined by $\Omega_c = \mathds{1}\left(y(\omega,c) = 1\right) \times \mathds{1}\left( \hat{y}(\omega,c) < \tau \right)$}.
\textcolor{red}{For unlabelled data in $\mathcal{U}$, $\Omega_c$ is defined as}
\begin{equation}
\textcolor{red}{
\begin{split}
\Omega_c = & \mathds{1}\left(\hat{y}(\omega,c) < \tau \right) \times \mathds{1}\left(\max_{\hat{c} \in \{1,\ldots,C\}} \bar{y}_{r}(\omega,\hat{c}) > \tau_w \right) \times \\
& \mathds{1}\left(\arg\max_{\hat{c} \in \{1,\ldots,C\}} \bar{y}_{r}(\omega,\hat{c}) = c \right),
\end{split}
}
\end{equation}
\textcolor{red}{where $\tau_w \in [0,1]$ is used to filter out extremely unreliable pseudo-labels, with $\tau_w < \tau$.}


\textcolor{red}{\textbf{Negative Set.}}
\textcolor{red}{The negative set for class $c$ is defined as}
\begin{equation}
\label{eq:negative_set}
\textcolor{red}{
\mathcal{R}^{-}_c  = \left\{ r(\omega,:) \mid \omega \in \Omega^{-}_c, \, r = \mathsf{Conv}_{1\times 1\times 1}\left(\mathsf{Conv}_{3\times 3\times 3}(f_{4})\right) \right\},
}
\end{equation}
\textcolor{red}{where for labelled data in $\mathcal{L}$, $\Omega^{-}_c = \mathds{1}\left(y(\omega,c) \ne 1 \right)$,}
\textcolor{red}{For unlabelled data in $\mathcal{U}$, $\Omega^{-}_c$ is defined by}
\begin{equation}
\label{eq:negative_Omega_unlabelled}
\textcolor{red}{
\begin{split}
\Omega^{-}_c = \mathds{1}\left(\max_{\hat{c} \in \{1,\ldots,C\}} \bar{y}_{r}(\omega,\hat{c}) > \tau_w \right) \times \mathds{1}\left(\arg\max_{\hat{c} \in \{1,\ldots,C\}} \bar{y}_{r}(\omega,\hat{c}) \ne c\right).
\end{split}
}
\end{equation}

\begin{figure}[!t]
\centering
\includegraphics[width=3.5in]{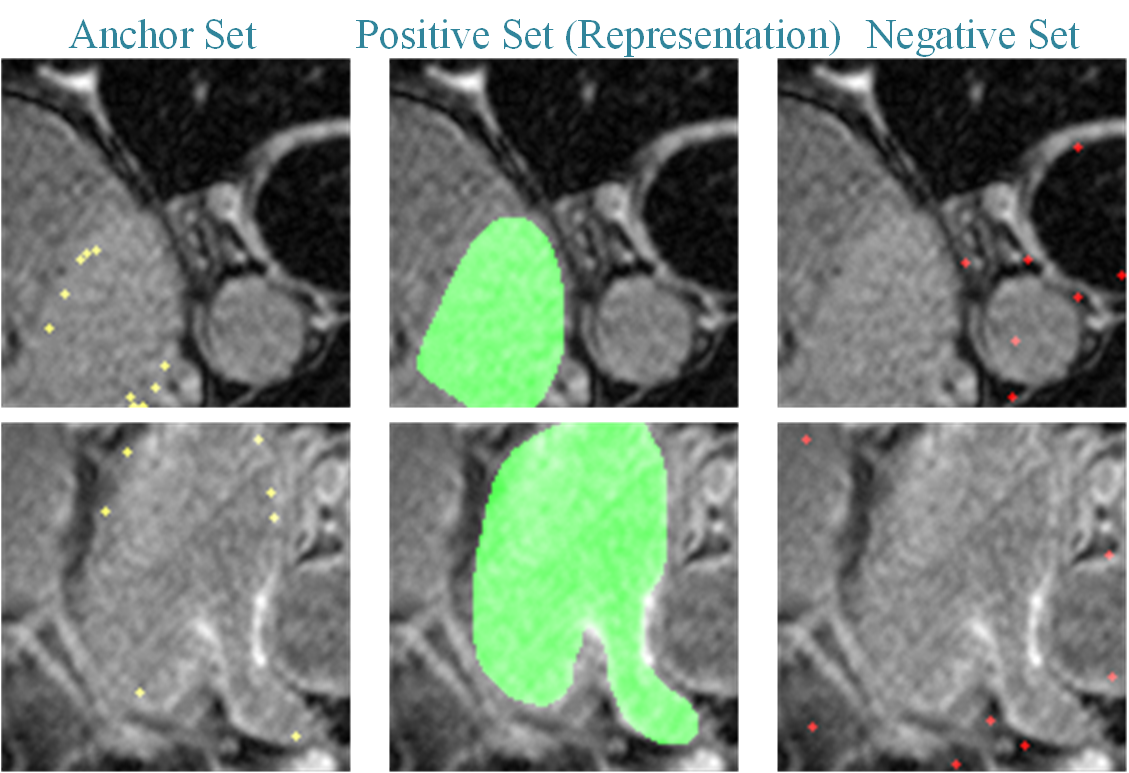}
\caption{\textcolor{red}{Examples of the anchor set and the corresponding negative and positive (representation part) sets on the Left Atrial dataset.}}
\label{fig:positive_negative}
\end{figure}

~\textcolor{red}{\textbf{Positive Set.}} For the class $c$ positive set, we need to consider that the uncertain regions can be challenging for the student model because their representations may deviate significantly from the class prototypes.
Therefore, the samples in the positive set should contain the key characteristics of the class, but these samples should not be too far from the uncertain representations to help the learning process--we refer to these positive samples as unassertive.
Based on this argument, we propose a positive set containing samples formed with the combination of the learned class prototypes $P$ 
and the mean representation of the current class. 
More specifically, the class-specific positive sample is defined by 
\begin{equation}
\textcolor{red}{
\mathcal{R}_c^{+} = \left \{r \Big | r = \frac{r_m + \xi \times \mathsf{mean}(P^c)}{1+\xi}, \xi \sim \mathbb{U}(0,1) \right \}, 
\label{eq:collaborative_center}
}
\end{equation}
\textcolor{red}{where $\mathbb{U}(0,1)$ generates a uniformly distributed random number in $(0,1]$,}
$P^c \in \mathbb{R}^{R \times F}$ denotes the learned prototypes for class $c$ with $\mathsf{mean}(\cdot)$ being the mean operator to produce the class-specific representation of the prototypes of class $c$, and $r_m$ is the mean of the features computed with $r = \mathsf{Conv}_{1\times 1\times 1}(\mathsf{Conv}_{3\times 3\times 3}(f_{4}))$ from labelled data $(x,y) \in \mathcal{L}$, where $y(\omega,c) = 1$, and from unlabelled data $x \in \mathcal{U}$, where $\max_{\hat{c} \in \{1,...,C\} }(\bar{y}_{r}(\omega,\hat{c})) > \tau_w$, and $\arg\max_{\hat{c} \in \{1,...,C\} }\bar{y}_{r}(\omega,\hat{c}) = c$. ~\textcolor{red}{Fig.~\ref{fig:positive_negative} shows some examples of the anchor set and the corresponding negative and positive (representation part) sets on the LA dataset.}

~\textcolor{red}{After obtaining the anchor, negative and positive sets,} the contrastive loss of the proposed CPS is formulated as
\begin{equation}
\label{eq:cp}
\ell_{cp}(\mathcal{L},\mathcal{U},\theta) = - \sum_{c \in \{1,...,C\}} \sum_{r_c \in \mathcal{R}_c} \log\left[\frac{e^{cos(r_c,r^+_c) / t}}{e^{cos(r_c,r^+_c) / t}+\sum _{r^-_c \in \mathcal{R}_c^{-}}{e^{cos(r_c,r^-_c) /t}}}\right].
\end{equation}
where 
$t$ represents the temperature parameter of the InfoNCE loss.

\subsection{Our Holistic Training Objective}\label{sec:Holistic_Training_Objective}
To take into account CRLN's learning of prototypes, DIM's learning of cross-attention block parameters, and CPS's representation learning, we extend the overall loss in~\eqref{eq:training_loss} as:
\begin{equation}
\begin{split}
\ell(\mathcal{L},\mathcal{U},\theta,P)  = & \frac{1}{L} \sum_{(x,y)\in\mathcal{L}} \ell_s \left (f_{\theta_{sd}}\left (f_{\theta_{se}}(x) \right ),y \right )+\ell_s \left (\mathscr{M}(x),y \right ) + \\ & \frac{1}{U} \sum_{(x)\in\mathcal{U}} \ell_u \left (f_{\theta_{sd}} \left (f_{\theta_{se}}(\mathscr{A}_s(x)) \right ),\bar{y}_{r} \right ) + \\
&\ell_{cp}(\mathcal{L},\mathcal{U},\theta),
\end{split}
\label{eq:training_loss1}
\end{equation}
where $\theta=\{\theta_{sd}, \theta_{se}, \theta_{td}, \theta_{te}, \theta_{sdim}, \theta_{tdim}\}$ (with $\theta_{sdim}, \theta_{tdim}$ representing the student's and teacher's DIM parameters defined in~\eqref{eq:dim}) and $P=\{P_t,P_s\}$ (with $P_t,P_s$ denoting the student's and teacher's prototypes). 
We also learn DIM's correction parameter $\mu$ in~\eqref{eq:rectification_student} with the following loss function:
\begin{equation}
\label{eq:training_loss2}
\ell_r(\mathcal{L},\mu)=\frac{1}{L} \sum_{(x,y)\in\mathcal{L}} \ell_s(\hat{y}_r,y),
\end{equation}
with $\hat{y}_r=\hat{y} + (1-\mu)\times \mathscr{M}(x)$.
The training alternates the minimisation of the losses in~\eqref{eq:training_loss1} and~\eqref{eq:training_loss2} and the Exponential Moving Average (EMA) to update the teacher's parameters, as described in Algorithm~\ref{alg:network_pseudocode}.



\begin{algorithm}[t]
\caption{Pseudo-code of the proposed method 
} 
\label{alg:network_pseudocode}
\begin{algorithmic}[1]
\State \textcolor{gray}{ \# Teacher model parameters: $\theta _{te}$ and $\theta _{td}$, DIM parameters (rectification stage): $\theta _{tdim}$, class prototypes: $P_t$}
\State \textcolor{gray}{ \# Student model parameters: $\theta _{se}$ and $\theta _{sd}$, DIM parameters (learning stage): $\theta _{sdim}$, class prototypes: $P_s$ } 
\State Initialise the model parameters 
\State  for {$index$ \textbf{in} $max iterations$:}
\State \hspace{0.3cm} for {$(x_l, y_l)$ \textbf{in} $\mathcal{L}$ and for $(x_u)$ \textbf{in} $\mathcal{U}$:}
\State \hspace{0.5cm} \textcolor{gray}{ \# Teacher model generates and rectifies pseudo-labels}
\State \hspace{0.5cm}  $\bar{y}_u \leftarrow f_{\theta _{td}}(f_{\theta_{te}}(\mathscr{A}_w(x_u)))$
\State \hspace{0.5cm} if $index$ \textgreater $S$:
\State \hspace{1cm} $\mathscr{M}(x_u) \leftarrow f_{\theta _{tdim}}(x_u)$  
\State \hspace{1cm} $\bar{y}_{r} = \bar{y} + (1-\mu)\cdot \mathscr{M}(x_u)$
\State \hspace{0.5cm} else:
\State \hspace{1cm} $\bar{y}_{r}=\bar{y}$
\State \hspace{0.5cm} \textcolor{gray}{ \# Train the student's model and prototypes}
\State \hspace{0.5cm} $\hat{y}_u,\hat{y}_l \leftarrow f_{\theta _{sd}}\left(f_{\theta _{se}}(\mathscr{A}_s(x_u))\right),f_{\theta _{sd}}\left (f_{\theta _{se}}(x_l) \right)$  
\State \hspace{0.5cm} $\mathscr{M}(x_l) \leftarrow f_{\theta _{sdim}}(x_l)$ 
\State \hspace{0.5cm} Update $\theta_{se}$, $\theta_{sd}$, $\theta_{sdim}$ and $P_s$ by minimising $\ell(\mathcal{L},\mathcal{U},\theta,P)$ in~\eqref{eq:training_loss1}
\State \hspace{0.5cm} \textcolor{gray}{ \# Learn $\mu$ in ~\eqref{eq:rectification_student}}
\State \hspace{0.5cm} Update $\mu$ by minimising $\ell_r(\mathcal{L},\theta_r)$ in~\eqref{eq:training_loss2}
\State \hspace{0.5cm} \textcolor{gray}{ \# Update the teacher's model and  prototypes}\State \hspace{0.5cm} Update the teacher's parameters $\theta_{te}$, $\theta_{td}$, $\theta_{tdim}$ and $P_t$ using the EMA of the student's parameters $\theta_{se}$, $\theta_{sd}$, $\theta_{sdim}$ and $P_s$.
\end{algorithmic}
\end{algorithm}

\section{Experiments}\label{sec:Experiments}

\subsection{Datasets}

The effectiveness of the proposed method is assessed on three public 3D medical image datasets with diverse tissue types, including the Left Atrium (LA)\citep{xiong2021global}, Pancreas-CT\citep{clark2013cancer}, and Brain Tumour Segmentation 2019 (BraTS19)\citep{menze2014multimodal} datasets. \par 

\textbf{LA dataset.} The LA dataset \citep{xiong2021global} is a standard 3D medical image semi-supervised segmentation benchmark, consisting of 100 3D MRI volumes with a fixed resolution of 0.625 $\times$ 0.625 $\times$ 0.625 mm. Following former works \citep{liu2022translation,yu2019uncertainty}, all volumes are cropped to 112 $\times$ 112 $\times$ 80, centred on the heart region. 80 volumes are used as the training dataset, while the remaining 20 volumes are allocated for validation.\par
\textbf{Pancreas-CT dataset.} The Pancreas-CT dataset \citep{clark2013cancer} is collected from 53 male and 27 female subjects at the National Institutes of Health Clinical Centre. It contains 82 3D contrast-enhanced CT scans. The size of each scan is 512 $\times$ 512, but the thickness varies from 1.5 to 2.5 mm. We adopt the same pre-processing methods as prior studies \citep{liu2022translation,wu2022mutual}, including clipping the voxel values to the range of [-125, 275] Hounsfield Units, re-sampling the data into an isotropic resolution of 1.0 $\times$ 1.0 $\times$ 1.0 mm and cropping data into 96 $\times$ 96 $\times$ 96. Then, 62 scans are utilised to train the model, and 20 scans are used for validation. \par 
\textbf{BraTS19 dataset.} The BraTS19 dataset \citep{menze2014multimodal} is composed of 335 labelled brain tumour MRIs with a size of 240 $\times$ 240 $\times$ 155. There are four MRI scans for each sample, including T1-weighted (T1), T1-weighted with contrast enhancement (T1-ce), T2-weighted (T2), and T2 fluid-attenuated inversion recovery (FLAIR). Consistent with \citet{chen2019multi}, only the FLAIR sequences are employed in the semi-supervised segmentation task. Moreover, we apply the identical dataset split, allocating 250 samples for training, 25 for validation, and 60 for testing. In the pre-processing stage, the training samples are \textcolor{red}{randomly} cropped into 96 $\times$ 96 $\times$ 96, in line with \citet{liu2022translation}.

\begin{table*}[!t]
{\caption{Comparative results on the Left Atrium dataset based on the partition protocols of 8 labelled data. The "Best" column represents the best checkpoint evaluation protocol ($\checkmark$: best checkpoint, $\times$: last checkpoint). \textcolor{red}{$\textsuperscript{*}$ denotes that the results (mean $\pm$ standard deviation) are calculated using the bootstrapping method.}
\label{tab:table1}}
\centering
\resizebox{\textwidth}{!}{
\renewcommand{\arraystretch}{0.85}
\begin{tabular}{c c c c l l l l}
\toprule[1pt]
\multirow{2}{*}{Left Atrium} & \multirow{2}{*}{{Best}} & \multicolumn{2}{c}{Scan Used}&\multirow{2}{*} {Dice(\%)}&\multirow{2}{*} {Jaccard(\%)}&\multirow{2}{*}{ASD(Voxel)}&\multirow{2}{*}{95HD(Voxel)}\\
\cline{3-4}
& & labelled & unlabelled\\
\hline
{UA-MT}\citep{yu2019uncertainty} & {$\times$} & {8} & {72} & {84.25} & {73.48} & {3.36} & {13.84}\\
\textcolor{red}{LG-ER \citep{hang2020local}\textsuperscript{*}} & \textcolor{red}{$\times$} & \textcolor{red}{8} & \textcolor{red}{72} & \textcolor{red}{85.24\scriptsize{$\pm$1.23}} & \textcolor{red}{74.69\scriptsize{$\pm$1.75}} & \textcolor{red}{3.77\scriptsize{$\pm$0.78}} & \textcolor{red}{14.99\scriptsize{$\pm$2.93}} \\
{DUWM}\citep{wang2020double}&	{$\times$}&	{8}&{72}&	{85.91}&{75.75}&{3.31}&	{12.67}\\
{URPC}\citep{luo2021efficient} &{$\times$}&{8}&{72}&	85.01&	74.36&	3.96&	15.37\\
\textcolor{red}{SASSNet\citep{li2020shape}\textsuperscript{*}}&\textcolor{red}{$\times$}&\textcolor{red}{8}&\textcolor{red}{72}&	\textcolor{red}{86.65\scriptsize$\pm$0.94}&	\textcolor{red}{76.69\scriptsize$\pm$1.42}&	\textcolor{red}{4.17\scriptsize$\pm$0.47}&\textcolor{red}{14.69\scriptsize$\pm$2.51}\\
TraCoCo\citep{liu2022translation} &{$\times$}&{8}&{72}&89.29&	80.82&	2.28& 6.92\\
ASE-Net\citep{lei2022semi} &{$\times$}&{8}&{72} &87.83&	78.45&2.17&	9.86\\
TAC\citep{chen2022semi} &{$\times$}&{8}&{72} &	{84.73}&	{74.38}&	{2.72}&	{11.45}\\
DCR\citep{LU2024107991}&{$\times$}&{8}&{72} &	{89.05}&	{80.27}&	{2.84}&	{9.72}\\
\textcolor{red}{CAC4SSL\citep{li2024contour}\textsuperscript{*}} & \textcolor{red}{$\times$} & \textcolor{red}{8} & \textcolor{red}{72} & \textcolor{red}{89.92\scriptsize$\pm$0.48} & \textcolor{red}{81.77\scriptsize$\pm$0.78} & \textcolor{red}{2.04\scriptsize$\pm$0.16} & \textcolor{red}{6.38\scriptsize$\pm$0.50} \\
\textcolor{red}{RCPS\citep{10273222}\textsuperscript{*}}&\textcolor{red}{{$\times$}}&\textcolor{red}{8}&\textcolor{red}{72} &\textcolor{red}{90.71\scriptsize$\pm$0.08}&\textcolor{red}{83.10\scriptsize$\pm$0.13}&\textcolor{red}{2.04\scriptsize$\pm$0.03}&\textcolor{red}{7.91\scriptsize$\pm$0.20}\\	
\rowcolor{gray!25} \textcolor{red}{\textbf{Ours}\textsuperscript{*}} &\textcolor{red}{$\times$}&\textcolor{red}{8}&\textcolor{red}{72}&\textcolor{red}{\textbf{91.74}\scriptsize$\pm$0.38}&\textcolor{red}{\textbf{84.79}\scriptsize$\pm$0.64}&\textcolor{red}{\textbf{1.46}\scriptsize$\pm$0.09}&\textcolor{red}{\textbf{4.60}\scriptsize$\pm$0.33}\\
\bottomrule[1pt]
\textcolor{red}{DTC\citep{luo2021semi}\textsuperscript{*}}&\textcolor{red}{$\checkmark$}&\textcolor{red}{8}&\textcolor{red}{72} &\textcolor{red}{87.25\scriptsize$\pm$1.28} &\textcolor{red}{77.80\scriptsize$\pm$1.83}&\textcolor{red}{2.44\scriptsize$\pm$0.31}&\textcolor{red}{8.52\scriptsize$\pm$1.15}\\
{SDC-SSL}\citep{lei2023shape}&{$\checkmark$} &{8} & {72} & {88.31} & {79.25} & {1.94} & {7.56}\\
\textcolor{red}{MC-Net+\citep{wu2022mutual}\textsuperscript{*}}&\textcolor{red}{$\checkmark$}&\textcolor{red}{8}&\textcolor{red}{72} &\textcolor{red}{88.76\scriptsize$\pm$0.73}&\textcolor{red}{79.93\scriptsize$\pm$1.16}&\textcolor{red}{1.93\scriptsize$\pm$0.16}&\textcolor{red}{8.16\scriptsize$\pm$0.92}\\
\textcolor{red}{CAML\citep{gao2023correlation}\textsuperscript{*}}&\textcolor{red}{$\checkmark$}&\textcolor{red}{8}&\textcolor{red}{72}&\textcolor{red}{89.34\scriptsize$\pm$0.55}&\textcolor{red}{80.84\scriptsize$\pm$0.89}&\textcolor{red}{2.15\scriptsize$\pm$0.40}&\textcolor{red}{10.37\scriptsize$\pm$2.44}\\
\textcolor{red}{BCP\citep{bai2023bidirectional}\textsuperscript{*}} & \textcolor{red}{$\checkmark$} & \textcolor{red}{8} & \textcolor{red}{72} & \textcolor{red}{89.83\scriptsize$\pm$0.53} & \textcolor{red}{81.64\scriptsize$\pm$0.87} & \textcolor{red}{1.84\scriptsize$\pm$0.15} & \textcolor{red}{6.92\scriptsize$\pm$0.66} \\
MLRPL\citep{su2024mutual}&{$\checkmark$}&{8}&{72}&89.86&81.68&	1.85&6.91\\
\textcolor{red}{TraCoCo\citep{liu2022translation}\textsuperscript{*}} & \textcolor{red}{$\checkmark$} & \textcolor{red}{8} & \textcolor{red}{72} & \textcolor{red}{89.78\scriptsize$\pm$0.71} & \textcolor{red}{81.58\scriptsize$\pm$1.13} & \textcolor{red}{1.98\scriptsize$\pm$0.21} & \textcolor{red}{6.59\scriptsize$\pm$0.64} \\
\rowcolor{gray!25} \textcolor{red}{\textbf{Ours}\textsuperscript{*}} & \textcolor{red}{$\checkmark$} & \textcolor{red}{8} & \textcolor{red}{72} & \textcolor{red}{\textbf{91.85\scriptsize$\pm$0.37}} & \textcolor{red}{\textbf{84.98\scriptsize$\pm$0.63}} & \textcolor{red}{\textbf{1.41\scriptsize$\pm$0.08}} & \textcolor{red}{\textbf{4.62\scriptsize$\pm$0.31}} \\
\bottomrule[1pt]
\end{tabular}
}}
\end{table*}

\begin{table*}[!t]
{\caption{Comparative results on the Left Atrium dataset based on the partition protocols of 16 labelled data. The "Best" column represents the best checkpoint evaluation protocol ($\checkmark$: best checkpoint, $\times$: last checkpoint).\textcolor{red}{$\textsuperscript{*}$ denotes that the results (mean $\pm$ standard deviation) are calculated using the bootstrapping method.}\label{tab:table2}}
\centering
\resizebox{\textwidth}{!}{
\renewcommand{\arraystretch}{0.85}
\begin{tabular}{c c c c l l l l}
\toprule[1pt]
\multirow{2}{*}{Left Atrium} & \multirow{2}{*}{{Best}} & \multicolumn{2}{c}{Scan Used}&\multirow{2}{*} {Dice(\%)}&\multirow{2}{*} {Jaccard(\%)}&\multirow{2}{*}{ASD(Voxel)}&\multirow{2}{*}{95HD(Voxel)}\\
\cline{3-4}
& & labelled & unlabelled\\
\hline
\textcolor{red}{UA-MT \citep{yu2019uncertainty}\textsuperscript{*}} & \textcolor{red}{$\times$} & \textcolor{red}{16} & \textcolor{red}{64} & \textcolor{red}{88.78\scriptsize$\pm$0.96} & \textcolor{red}{80.05\scriptsize$\pm$1.5} & \textcolor{red}{2.27\scriptsize$\pm$0.20} & \textcolor{red}{7.41\scriptsize$\pm$0.68} \\
MCF \citep{wang2023mcf} & $\times$ & 16 & 64 & 88.71 & 80.41 & 1.90 & 6.32 \\
URPC \citep{luo2021efficient} & $\times$ & 16 & 64 & 88.74 & 79.93 & 3.66 & 12.73 \\
\textcolor{red}{SASSNet \citep{li2020shape}\textsuperscript{*}} & \textcolor{red}{$\times$} & \textcolor{red}{16} & \textcolor{red}{64} & \textcolor{red}{89.02\scriptsize$\pm$0.94} & \textcolor{red}{80.45\scriptsize$\pm$1.45} & \textcolor{red}{2.98\scriptsize$\pm$0.55} & \textcolor{red}{8.83\scriptsize$\pm$1.61} \\
\textcolor{red}{LG-ER \citep{hang2020local}\textsuperscript{*}} & \textcolor{red}{$\times$} & \textcolor{red}{16} & \textcolor{red}{64} & \textcolor{red}{89.50\scriptsize$\pm$0.60} & \textcolor{red}{81.10\scriptsize$\pm$0.97} & \textcolor{red}{2.08\scriptsize$\pm$0.18} & \textcolor{red}{7.33\scriptsize$\pm$0.66} \\
DUWM \citep{wang2020double} & $\times$ & 16 & 64 & 89.65 & 81.35 & 2.03 & 7.04 \\
\textcolor{red}{CAC4SSL \citep{li2024contour}\textsuperscript{*}} & \textcolor{red}{$\times$} & \textcolor{red}{16} & \textcolor{red}{64} & \textcolor{red}{90.55\scriptsize$\pm$0.59} & \textcolor{red}{82.84\scriptsize$\pm$0.96} & \textcolor{red}{1.71\scriptsize$\pm$0.14} & \textcolor{red}{5.98\scriptsize$\pm$0.53} \\
TraCoCo \citep{liu2022translation} & $\times$ & 16 & 64 & 90.94 & 83.47 & 1.79 & 5.49 \\
ASE-Net \citep{lei2022semi} & $\times$ & 16 & 64 & 90.29 & 82.76 & 1.64 & 7.18 \\
TAC \citep{chen2022semi} & $\times$ & 16 & 64 & 87.75 & 78.60 & 2.04 & 9.45 \\
DCR \citep{LU2024107991} & $\times$ & 16 & 64 & 91.21 & 83.54 & 1.92 & 8.03 \\
\textcolor{red}{RCPS \citep{10273222}\textsuperscript{*}} & \textcolor{red}{$\times$} & \textcolor{red}{16} & \textcolor{red}{64} & \textcolor{red}{91.19\scriptsize$\pm$0.08} & \textcolor{red}{83.87\scriptsize$\pm$0.13} & \textcolor{red}{1.78\scriptsize$\pm$0.03} & \textcolor{red}{6.34\scriptsize$\pm$0.21} \\
\rowcolor{gray!25} \textcolor{red}{\textbf{Ours}\textsuperscript{*}} & \textcolor{red}{$\times$} & \textcolor{red}{16} & \textcolor{red}{64} & \textcolor{red}{\textbf{91.88\scriptsize$\pm$0.43}} & \textcolor{red}{\textbf{85.05\scriptsize$\pm$0.72}} & \textcolor{red}{\textbf{1.46\scriptsize$\pm$0.08}} & \textcolor{red}{\textbf{4.51\scriptsize$\pm$0.32}} \\
\bottomrule[1pt]
\textcolor{red}{DTC \citep{luo2021semi}\textsuperscript{*}} & \textcolor{red}{$\checkmark$} & \textcolor{red}{16} & \textcolor{red}{64} & \textcolor{red}{89.40\scriptsize$\pm$0.56} & \textcolor{red}{80.94\scriptsize$\pm$0.91} & \textcolor{red}{1.91\scriptsize$\pm$0.17} & \textcolor{red}{7.76\scriptsize$\pm$0.73} \\
{SDC-SSL}\citep{lei2023shape}&{$\checkmark$} &{16} & {64} & {90.44} & {82.73} & {1.75} & {6.02}\\
\textcolor{red}{CAML \citep{gao2023correlation}\textsuperscript{*}} & \textcolor{red}{$\checkmark$} & \textcolor{red}{16} & \textcolor{red}{64} & \textcolor{red}{90.65\scriptsize$\pm$0.50} & \textcolor{red}{82.98\scriptsize$\pm$0.83} & \textcolor{red}{1.62\scriptsize$\pm$0.14} & \textcolor{red}{6.12\scriptsize$\pm$0.50} \\
\textcolor{red}{MC-Net+ \citep{wu2022mutual}\textsuperscript{*}} & \textcolor{red}{$\checkmark$} & \textcolor{red}{16} & \textcolor{red}{64} & \textcolor{red}{90.99\scriptsize$\pm$0.45} & \textcolor{red}{83.55\scriptsize$\pm$0.75} & \textcolor{red}{1.71\scriptsize$\pm$0.15} & \textcolor{red}{5.84\scriptsize$\pm$0.52} \\
MLRPL\citep{su2024mutual}&{$\checkmark$}&{16}&{64}&91.02&83.62&	1.66&5.78\\
\textcolor{red}{TraCoCo \citep{liu2022translation}\textsuperscript{*}} & \textcolor{red}{$\checkmark$} & \textcolor{red}{16} & \textcolor{red}{64} & \textcolor{red}{91.44\scriptsize$\pm$0.41} & \textcolor{red}{84.30\scriptsize$\pm$0.69} & \textcolor{red}{1.79\scriptsize$\pm$0.16} & \textcolor{red}{5.62\scriptsize$\pm$0.52} \\
\rowcolor{gray!25} \textcolor{red}{\textbf{Ours}\textsuperscript{*}} & \textcolor{red}{$\checkmark$} & \textcolor{red}{16} & \textcolor{red}{64} & \textcolor{red}{\textbf{91.95\scriptsize$\pm$0.38}} & \textcolor{red}{\textbf{85.15\scriptsize$\pm$0.64}} & \textcolor{red}{\textbf{1.45\scriptsize$\pm$0.08}} & \textcolor{red}{\textbf{4.33\scriptsize$\pm$0.28}} \\
\bottomrule[1pt]
\end{tabular}
}}
\end{table*}

\begin{table*}[!t]
{\caption{Comparative results on the Pancreas-CT dataset based on the partition protocols of 6 and 12 labelled data. The "Best" column represents the best checkpoint evaluation protocol ($\checkmark$: best checkpoint, $\times$: last checkpoint).\textcolor{red}{$\textsuperscript{*}$ denotes that the results (mean $\pm$ standard deviation) are calculated using the bootstrapping method.}\label{tab:table3}}
\centering
\resizebox{\textwidth}{!}{
\renewcommand{\arraystretch}{0.85}
\begin{tabular}{c c c c l l l l}
\toprule[1pt]
\multirow{2}{*}{Pancreas-CT} & \multirow{2}{*}{{Best}} & \multicolumn{2}{c}{Scan Used}&\multirow{2}{*} {Dice(\%)}&\multirow{2}{*} {Jaccard(\%)}&\multirow{2}{*}{ASD(Voxel)}&\multirow{2}{*}{95HD(Voxel)}\\
\cline{3-4}
& & labelled & unlabelled\\
\hline
{UA-MT}\citep{yu2019uncertainty} & {$\times$} & {6} & {56} & {66.44} & {52.02} & {3.03} & {17.04}\\
{SASSNet}\citep{li2020shape} &	{$\times$}&	{6} & {56}&	{68.97}&	{54.29}&	{1.96}&	{18.83}\\
{URPC}\citep{luo2021efficient}&{$\times$}&{6}&	{56}&73.53&	59.44&	7.85&	22.57\\
MCCauSSL\citep{miao2023caussl}&{$\times$} &{6}&	{56}&	72.89 &	58.06 &	4.37&14.19 \\
\textcolor{red}{CAC4SSL \citep{li2024contour}\textsuperscript{*}} & \textcolor{red}{$\times$} & \textcolor{red}{6} & \textcolor{red}{56} & \textcolor{red}{74.19\scriptsize$\pm$1.94} & \textcolor{red}{59.82\scriptsize$\pm$2.35} & \textcolor{red}{3.01\scriptsize$\pm$0.81} & \textcolor{red}{15.36\scriptsize$\pm$3.02} \\
\textcolor{red}{RCPS \citep{10273222}\textsuperscript{*}} & \textcolor{red}{$\times$} & \textcolor{red}{6} & \textcolor{red}{56} & \textcolor{red}{76.81\scriptsize$\pm$0.12} & \textcolor{red}{63.07\scriptsize$\pm$0.15} & \textcolor{red}{2.90\scriptsize$\pm$0.06} & \textcolor{red}{15.43\scriptsize$\pm$0.49} \\
\textcolor{red}{TraCoCo \citep{liu2022translation}\textsuperscript{*}} & \textcolor{red}{$\times$} & \textcolor{red}{6} & \textcolor{red}{56} & \textcolor{red}{79.10\scriptsize$\pm$1.32} & \textcolor{red}{65.90\scriptsize$\pm$1.76} & \textcolor{red}{2.58\scriptsize$\pm$0.57} & \textcolor{red}{8.42\scriptsize$\pm$1.28} \\
\rowcolor{gray!25} \textcolor{red}{\textbf{Ours}\textsuperscript{*}} & \textcolor{red}{$\times$} & \textcolor{red}{6} & \textcolor{red}{56} & \textcolor{red}{\textbf{79.17\scriptsize$\pm$1.82}} & \textcolor{red}{\textbf{66.32\scriptsize$\pm$2.33}} & \textcolor{red}{\textbf{1.94\scriptsize$\pm$0.52}} & \textcolor{red}{\textbf{7.27\scriptsize$\pm$1.12}} \\	
\bottomrule[1pt]
\textcolor{red}{MC-Net+ \citep{wu2022mutual}\textsuperscript{*}} & \textcolor{red}{$\checkmark$} & \textcolor{red}{6} & \textcolor{red}{56} & \textcolor{red}{69.44\scriptsize$\pm$3.05} & \textcolor{red}{55.16\scriptsize$\pm$3.20} & \textcolor{red}{3.74\scriptsize$\pm$0.85} & \textcolor{red}{16.72\scriptsize$\pm$3.38} \\
MLRPL \citep{su2024mutual} & {$\checkmark$} & 6 & 56 & 75.93 & 62.12 & \textbf{1.54} & 9.07 \\
\rowcolor{gray!25} \textcolor{red}{\textbf{Ours}\textsuperscript{*}} & \textcolor{red}{$\checkmark$} & \textcolor{red}{6} & \textcolor{red}{56} & \textcolor{red}{\textbf{81.24\scriptsize$\pm$1.23}} & \textcolor{red}{\textbf{68.84\scriptsize$\pm$1.71}} & \textcolor{red}{3.15\scriptsize$\pm$0.81} & \textcolor{red}{\textbf{8.38\scriptsize$\pm$2.42}} \\	
\bottomrule[1pt]
{UA-MT}\citep{yu2019uncertainty} & {$\times$} & {12} & {50} & {76.10} & {62.62} & {2.43} & {10.84}\\
{SASSNet}\citep{li2020shape} &	{$\times$}&	{12} & {50}&	{76.39}&	{63.17}&	{1.42}&	{11.06}\\
MCF \citep{wang2023mcf} & $\times$ & 12 & 50 & 75.00 & 61.27 & 3.27 & 11.59 \\
DCR\citep{LU2024107991}&{$\times$}&{12}&{50} &	{79.84}&	{67.72}&	{1.56}&	{7.23}\\		
{URPC}\citep{luo2021efficient} &{$\times$}&{12} & {50}&80.02&	67.30&	1.98&	8.51\\
MCCauSSL\citep{miao2023caussl}&{$\times$} &{12} & {50}&80.92& 68.26&	1.53&	8.11\\
\textcolor{red}{TraCoCo \citep{liu2022translation}\textsuperscript{*}} & \textcolor{red}{$\times$} & \textcolor{red}{12} & \textcolor{red}{50} & \textcolor{red}{81.67\scriptsize$\pm$1.09} & \textcolor{red}{69.40\scriptsize$\pm$1.54} & \textcolor{red}{1.47\scriptsize$\pm$0.23} & \textcolor{red}{5.68\scriptsize$\pm$0.69} \\
\textcolor{red}{CAC4SSL\citep{li2024contour}\textsuperscript{*}} & $\times$ & 12 & 50 & \textcolor{red}{80.43\scriptsize$\pm$1.45} & \textcolor{red}{67.85\scriptsize$\pm$1.94} & \textcolor{red}{7.25\scriptsize$\pm$1.21} & \textcolor{red}{\textbf{1.25\scriptsize$\pm$0.08}} \\
\textcolor{red}{RCPS \citep{10273222}\textsuperscript{*}} & \textcolor{red}{$\times$} & \textcolor{red}{12} & \textcolor{red}{50} & \textcolor{red}{81.62\scriptsize$\pm$0.17} & \textcolor{red}{69.25\scriptsize$\pm$0.23} & \textcolor{red}{2.00\scriptsize$\pm$0.05} & \textcolor{red}{7.20\scriptsize$\pm$0.31} \\
\rowcolor{gray!25} \textcolor{red}{\textbf{Ours}\textsuperscript{*}} & \textcolor{red}{$\times$} & \textcolor{red}{12} & \textcolor{red}{50} & \textbf{\textcolor{red}{82.76\scriptsize$\pm$1.01}} & \textbf{\textcolor{red}{70.90\scriptsize$\pm$1.45}} & \textbf{\textcolor{red}{1.35\scriptsize$\pm$0.19}} &\textcolor{red}{ 4.81\scriptsize$\pm$0.60} \\
\bottomrule[1pt]
{DTC}\citep{luo2021semi}& {$\checkmark$}&	{12} & {50}&	{78.27}&{64.75}&{2.25}&	{8.36}\\
\textcolor{red}{MC-Net+ \citep{wu2022mutual}\textsuperscript{*}} & \textcolor{red}{$\checkmark$} & \textcolor{red}{12} & \textcolor{red}{50} & \textcolor{red}{78.95\scriptsize$\pm$2.26} & \textcolor{red}{66.37\scriptsize$\pm$2.70} & \textcolor{red}{1.79\scriptsize$\pm$0.55} & \textcolor{red}{8.78\scriptsize$\pm$2.14} \\
MLRPL\citep{su2024mutual} & $\checkmark$ & 12 & 50 & 81.53 & 69.35 & {\textbf{1.33}} & 6.81 \\
BCP\citep{bai2023bidirectional} & $\checkmark$ & 12 & 50 & 82.91 & 70.97 & 2.25 & 6.43 \\
\textcolor{red}{TraCoCo \citep{liu2022translation}\textsuperscript{*}} & \textcolor{red}{$\checkmark$} & \textcolor{red}{12} & \textcolor{red}{50} & \textcolor{red}{83.31\scriptsize$\pm$0.84} & \textcolor{red}{71.64\scriptsize$\pm$1.23} & \textcolor{red}{1.70\scriptsize$\pm$0.37} & \textcolor{red}{7.02\scriptsize$\pm$2.29} \\
\rowcolor{gray!25} \textcolor{red}{\textbf{Ours}\textsuperscript{*}} & \textcolor{red}{$\checkmark$} & \textcolor{red}{12} & \textcolor{red}{50} & \textcolor{red}{\textbf{83.52\scriptsize$\pm$1.03}} & \textcolor{red}{\textbf{71.99\scriptsize$\pm$1.47}} & \textcolor{red}{1.46\scriptsize$\pm$0.24} & \textcolor{red}{\textbf{4.70\scriptsize$\pm$0.62}} \\
\bottomrule[1pt]
\end{tabular}
}}
\end{table*}

\begin{table*}[!t]
{\caption{Comparative results on the BRaTS2019 dataset based on the partition protocols of 25 and 50 labelled data.\textcolor{red}{$\textsuperscript{*}$ denotes that the results (mean $\pm$ standard deviation) are calculated using the bootstrapping method.}\label{tab:table4}.}
\centering
\resizebox{\textwidth}{!}{
\renewcommand{\arraystretch}{0.85}
\begin{tabular}{c c c l l l l}
\toprule[1pt]
\multirow{2}{*}{Left Atrium} & \multicolumn{2}{c}{Scan Used}&\multirow{2}{*} {Dice(\%)}&\multirow{2}{*} {Jaccard(\%)}&\multirow{2}{*}{ASD(Voxel)}&\multirow{2}{*}{95HD(Voxel)}\\
\cline{2-3}
&  labelled & unlabelled\\
\hline
{UA-MT}\citep{yu2019uncertainty}  & {25} & {225} & {84.64} & {74.76} & {2.36} & {10.47}\\
{SASSNet}\citep{li2020shape}  & {25} & {225} & {84.73} & {74.89} & {2.44} & {9.88}\\
{LG-ER}\citep{hang2020local}  & {25} & {225} & {84.75} & {74.97} & {2.21} & {9.56}\\
{URPC}\citep{luo2021efficient}  & {25} & {225} & {84.53} & {74.60} & {2.55} & {9.79}\\
{MC-Net+}\citep{wu2022mutual}  & {25} & {225} & {84.96} & {75.14} & {2.36} & {9.45}\\
{LG-ER}\citep{hang2020local}  & {25} & {225} & {84.75} & {74.97} & {2.21} & {9.56}\\
{SDC-SSL}\citep{lei2023shape}  &{25} & {225} & {84.76} & {75.11} & {1.95} & {11.29}\\
{MCCauSSL}\citep{miao2023caussl}  & {25} & {225} & {83.54} & {73.46} & {1.98} & {12.53}\\
MLRPL\citep{su2024mutual}&{25}&{225}&84.29&74.74&2.55&9.57\\
{BCP}\citep{bai2023bidirectional}  & {25} & {225} & {85.14} & {76.01} & {2.88} & {9.89}\\
\textcolor{red}{TraCoCo \citep{liu2022translation}\textsuperscript{*}} & \textcolor{red}{25} & \textcolor{red}{225} & \textcolor{red}{85.79\scriptsize$\pm$1.57} & \textcolor{red}{76.49\scriptsize$\pm$2.15} & \textcolor{red}{1.72\scriptsize$\pm$0.29} & \textcolor{red}{7.03\scriptsize$\pm$1.07} \\
\rowcolor{gray!25} \textcolor{red}{\textbf{Ours}\textsuperscript{*}} & \textcolor{red}{25} & \textcolor{red}{225} & \textcolor{red}{\textbf{86.73\scriptsize$\pm$1.46}} & \textcolor{red}{\textbf{77.85\scriptsize$\pm$2.02}} & \textcolor{red}{\textbf{1.67\scriptsize$\pm$0.30}} & \textcolor{red}{\textbf{7.15\scriptsize$\pm$1.12}} \\
\bottomrule[1pt]
{UA-MT}\citep{yu2019uncertainty}   & {50} & {200} & {85.32} & {75.93} & {1.98} & {8.68}\\
{SASSNet}\citep{li2020shape}  & {50} & {200}& {85.64} & {76.33} & {2.04} & {9.17}\\
{LG-ER}\citep{hang2020local}  & {50} & {200} & {85.67} & {76.36} & {1.99} & {8.92}\\
{URPC}\citep{luo2021efficient}  &  {50} & {200} & {85.38} & {76.14} & {1.87} & {8.36}\\
{SDC-SSL}\citep{lei2023shape}   &  {50} & {200} & {85.45} & {76.23} & {1.96} & {7.61}\\
MLRPL\citep{su2024mutual}&{50}&{200}&85.47&76.32&2.00&7.76\\
{MC-Net+}\citep{wu2022mutual}   &  {50} & {200} & {86.02} & {76.98} & {1.98} & {8.74}\\
{BCP}\citep{bai2023bidirectional}   &  {50} & {200} & {86.13} & {77.24} & {2.06} & {8.99}\\
{TraCoCo}\citep{liu2022translation}  & {50} & {200} & {86.69} & {77.69} & {1.93} & {8.04}\\
\rowcolor{gray!25} \textcolor{red}{\textbf{Ours}\textsuperscript{*}} & \textcolor{red}{50} & \textcolor{red}{200} & \textcolor{red}{\textbf{87.17\scriptsize$\pm$1.24}} & \textcolor{red}{\textbf{78.19\scriptsize$\pm$1.80}} & \textcolor{red}{\textbf{1.62\scriptsize$\pm$0.32}} & \textcolor{red}{\textbf{6.79\scriptsize$\pm$1.05}} \\
\bottomrule[1pt]
\end{tabular}
}}
\end{table*}

\begin{table*}[!t]
\caption{Ablation results on the Left Atrium and Pancreas-CT datasets. \textcolor{red}{"MR" denotes the mean representation and 'LP' is the learned prototype.}\label{tab:table5}}
\centering
\scalebox{0.98}{
\resizebox{\textwidth}{!}{
\renewcommand{\arraystretch}{0.85}
\begin{tabular}{c | c | c c| c c c| c c c c }
\toprule[1pt]
\textcolor{red}{Dataset} & \textcolor{red}{Baseline} & \textcolor{red}{CutMix+Random Noise} & \textcolor{red}{CRLN+DIM} & \textcolor{red}{MR} & \textcolor{red}{LP} & \textcolor{red}{CPS} & \textcolor{red}{Dice(\%)} & \textcolor{red}{Jaccard(\%)} & \textcolor{red}{ASD(Voxel)} & \textcolor{red}{95HD(Voxel)} \\
\hline
\multirow{5}{*}{\textcolor{red}{Left Atrium}}
& \textcolor{red}{\checkmark} &  & & & & & \textcolor{red}{84.19\scriptsize$\pm$0.10} & \textcolor{red}{72.97\scriptsize$\pm$0.59} & \textcolor{red}{3.95\scriptsize$\pm$0.61} & \textcolor{red}{13.12\scriptsize$\pm$2.58} \\
& \textcolor{red}{\checkmark} & \textcolor{red}{\checkmark} & & & & & \textcolor{red}{88.67\scriptsize$\pm$1.04} & \textcolor{red}{79.90\scriptsize$\pm$1.59} & \textcolor{red}{2.98\scriptsize$\pm$0.60} & \textcolor{red}{10.97\scriptsize$\pm$3.14} \\
& \textcolor{red}{\checkmark} & \textcolor{red}{\checkmark} & \textcolor{red}{\checkmark} & & & & \textcolor{red}{89.87\scriptsize$\pm$0.75} & \textcolor{red}{81.73\scriptsize$\pm$1.20} & \textcolor{red}{2.12\scriptsize$\pm$0.28} & \textcolor{red}{6.64\scriptsize$\pm$0.97} \\
& \textcolor{red}{\checkmark} & \textcolor{red}{\checkmark} & \textcolor{red}{\checkmark} & \textcolor{red}{\checkmark} & & & \textcolor{red}{91.63\scriptsize$\pm$0.37} & \textcolor{red}{84.59\scriptsize$\pm$0.63} & \textcolor{red}{1.49\scriptsize$\pm$0.09} & \textcolor{red}{4.77\scriptsize$\pm$0.32} \\
& \textcolor{red}{\checkmark} & \textcolor{red}{\checkmark} & \textcolor{red}{\checkmark} & & \textcolor{red}{\checkmark} & & \textcolor{red}{91.53\scriptsize$\pm$0.31} & \textcolor{red}{84.42\scriptsize$\pm$0.52} & \textcolor{red}{1.55\scriptsize$\pm$0.09} & \textcolor{red}{4.70\scriptsize$\pm$0.28} \\
& \textcolor{red}{\checkmark} & \textcolor{red}{\checkmark} & \textcolor{red}{\checkmark} & & & \textcolor{red}{\checkmark} & \textcolor{red}{\textbf{91.74}\scriptsize$\pm$0.38} & \textcolor{red}{\textbf{84.79}\scriptsize$\pm$0.64} & \textcolor{red}{\textbf{1.46}\scriptsize$\pm$0.09} & \textcolor{red}{\textbf{4.60}\scriptsize$\pm$0.33} \\
\hline
\multirow{5}{*}{\textcolor{red}{Pancreas-CT}}
& \textcolor{red}{\checkmark} & & & & & & \textcolor{red}{56.14\scriptsize$\pm$3.67} & \textcolor{red}{41.04\scriptsize$\pm$3.57} & \textcolor{red}{7.42\scriptsize$\pm$1.06} & \textcolor{red}{21.99\scriptsize$\pm$3.10} \\
& \textcolor{red}{\checkmark} & \textcolor{red}{\checkmark} & & & & & \textcolor{red}{76.73\scriptsize$\pm$2.12} & \textcolor{red}{63.28\scriptsize$\pm$2.59} & \textcolor{red}{2.59\scriptsize$\pm$0.70} & \textcolor{red}{9.89\scriptsize$\pm$2.52} \\
& \textcolor{red}{\checkmark} & \textcolor{red}{\checkmark} & \textcolor{red}{\checkmark} & & & & \textcolor{red}{78.58\scriptsize$\pm$1.47} & \textcolor{red}{65.27\scriptsize$\pm$1.97} & \textcolor{red}{3.19\scriptsize$\pm$0.66} & \textcolor{red}{9.83\scriptsize$\pm$2.46} \\
& \textcolor{red}{\checkmark} & \textcolor{red}{\checkmark} & \textcolor{red}{\checkmark} & \textcolor{red}{\checkmark} & & & \textcolor{red}{74.68\scriptsize$\pm$2.42} & \textcolor{red}{60.75\scriptsize$\pm$2.84} & \textcolor{red}{1.94\scriptsize$\pm$0.26} & \textcolor{red}{8.56\scriptsize$\pm$1.17} \\
& \textcolor{red}{\checkmark} & \textcolor{red}{\checkmark} & \textcolor{red}{\checkmark} & & \textcolor{red}{\checkmark} & & \textcolor{red}{73.90\scriptsize$\pm$1.82} & \textcolor{red}{59.31\scriptsize$\pm$2.24} & \textcolor{red}{2.00\scriptsize$\pm$0.29} & \textcolor{red}{9.67\scriptsize$\pm$1.26} \\
& \textcolor{red}{\checkmark} & \textcolor{red}{\checkmark} & \textcolor{red}{\checkmark} & & & \textcolor{red}{\checkmark} & \textcolor{red}{\textbf{79.17}\scriptsize$\pm$1.82} & \textcolor{red}{\textbf{66.32}\scriptsize$\pm$2.33} & \textcolor{red}{\textbf{1.94}\scriptsize$\pm$0.52} & \textcolor{red}{\textbf{7.27}\scriptsize$\pm$1.12} \\
\bottomrule[1pt]
\end{tabular}}
}
\end{table*}

\subsection{Implementation Details}
\textcolor{red}{\textbf{Training and Testing Configurations.}} To fairly compare with prior works, the same VNet \citep{milletari2016v} is adopted as our backbone architecture in the LA and Pancreas-CT datasets, and the same 3D-UNet \citep{cciccek20163d} is employed for the BraTS19 dataset. The experiments are conducted on the Ubuntu system with NVIDIA GeForce RTX 3090 GPU. Training relies on an SGD optimiser with momentum 0.9, weight decay 5$e$-4, and initial learning rate 2.5$e$-3 that follows a polynomial learning rate scheduler. The batch size is set to 4, consisting of 2 labelled samples and 2 unlabelled samples. In ~\eqref{eq:cp}, the temperature parameter $t=0.5$, with 512 negative samples drawn from $\mathcal{R}_c^{-}$ and 256 anchors drawn from $\mathcal{R}_c$, following ~\citet{liu2021bootstrapping}. 
We set the number of iterations to start the rectification $S$ at $800$ and the control parameter $\xi$ at $0.6$ in ~\eqref{eq:collaborative_center}.
We provide an ablation study about $S$ and $\xi$ in Sec.~\ref{sec:ablation_s} and Sec.~\ref{sec:ablation_xi}, respectively.
For the semi-supervised learning settings, we adopt the 10\% and 20\% partition protocols. 
For testing, we follow the same sliding window technique used in \citet{yu2019uncertainty} and \citet{wu2022mutual} to obtain the final prediction with the same resolution as the original test sample. Specially, the strides are set as 18 $\times$ 18 $\times$ 4 in the LA dataset, and as 16 $\times$ 16 $\times$ 16 in the Pancreas-CT and BraTS19 datasets. The code is available at \url{https://github.com/Yaan-Wang/CRLN.git}.

\textcolor{red}{\textbf{Compared Methods.}}
Nineteen advanced semi-supervised medical image segmentation methods are compared on the LA~\citep{xiong2021global}, Pancreas-CT~\citep{clark2013cancer}, and BraTS19~\citep{menze2014multimodal} datasets, including: UA-MT~\citep{yu2019uncertainty}, DUWM~\citep{wang2020double}, SASSNet~\citep{li2020shape}, DTC~\citep{luo2021semi}, LG-ER~\citep{hang2020local}, URPC~\citep{luo2021efficient}, TraCoCo~\citep{liu2022translation}, ASE-Net~\citep{lei2022semi}, TAC~\citep{chen2022semi}, MC-Net+~\citep{wu2022mutual}, SDC-SSL\citep{lei2023shape}, BCP~\citep{bai2023bidirectional}, MCF~\citep{wang2023mcf},  CAML~\citep{gao2023correlation}, MCCauSSL~\citep{miao2023caussl}, \textcolor{red}{RCPS ~\citep{10273222}}, CAC4SSL~\citep{li2024contour}, DCR~\citep{LU2024107991} and MLRPL~\citep{su2024mutual}. The results of these methods on the LA and Pancreas-CT datasets are sourced from the original papers, except URPC~\citep{luo2021efficient}, which is from TraCoCo~\citep{liu2022translation}. Additionally, the results on the BraTS19 dataset are sourced from TraCoCo~\citep{liu2022translation}, except SDC-SSL\citep{lei2023shape} and MLRPL~\citep{su2024mutual}, which are from the original papers. In addition, given the absence of a validation dataset for the LA and Pancreas-CT datasets, we categorise these comparison methods based on whether they utilise the results from the final checkpoint or the best checkpoint during the training process.


\textcolor{red}{\textbf{Evaluation Metrics.}}
Following the competing methods \citep{yu2019uncertainty,bai2023bidirectional}, four widely used measures are employed to evaluate the proposed approach, including Dice, Jaccard, the Average Surface Distance (ASD) and 95\% Hausdoff Distance (95HD). \textcolor{red}{We report the mean and standard deviation for each metric of the proposed method in all tables, using a bootstrapping method with 100 resamples. For the compared methods, we calculate and report the mean and standard deviation only for those with publicly available checkpoints to ensure their optimal performance.}

\subsection{Comparison with State-of-the-Art Approaches}

\textcolor{red}{\textbf{Segmentation Comparison on the LA Dataset.}} The comparative results under different protocols are reported in Tab.~\ref{tab:table1} and Tab.~\ref{tab:table2}. It can be seen that our proposed method provides a better result than other advanced approaches. For instance, with only 8 labelled samples available for training, \textcolor{red}{our Jaccard and 95HD results surpass the previous SOTA RCPS \citep{10273222} by 1.69\% and 3.31, respectively. With only 16 labelled training samples, the Dice of the last checkpoint and best checkpoint are raised to 91.88\% and 91.95\%, respectively.} The superiority of our approach stems from leveraging the labelled data as external priors to correct the pseudo-labels, which can indeed boost the overall segmentation quality. 

\textcolor{red}{\textbf{Segmentation Comparison on the Pancreas-CT Dataset.}} Different from other organs, the pancreas usually presents a more irregular and curved shape. This elongated and curvilinear morphology makes the segmentation of pancreas quite challenging. As illustrated in Tab.~\ref{tab:table3}, our proposed network achieves promising segmentation performance for all partition protocols. Compared to the strong baseline MC-Net+\citep{wu2022mutual}, \textcolor{red}{the proposed method demonstrates improvements of 11.8\% and 4.57\% on the Dice metric under the 10\% and 20\% partition protocols, respectively.} 

\textcolor{red}{\textbf{Segmentation Comparison on the BraTS19 Dataset.}}
Tab.~\ref{tab:table4} shows the results under different protocols. \textcolor{red}{The proposed approach consistently outperforms the current leading methods by a clear margin. Specifically, the Jaccard index is improved by 1.36\% and 0.5\% with 25 and 50 labelled samples for training, compared to the current SOTA TraCoCo \citep{liu2022translation}. These results demonstrate that the proposed method segments brain tumours more precisely than the compared approaches.}


\begin{figure}[!t]
\centering
\includegraphics[width=3.5in]{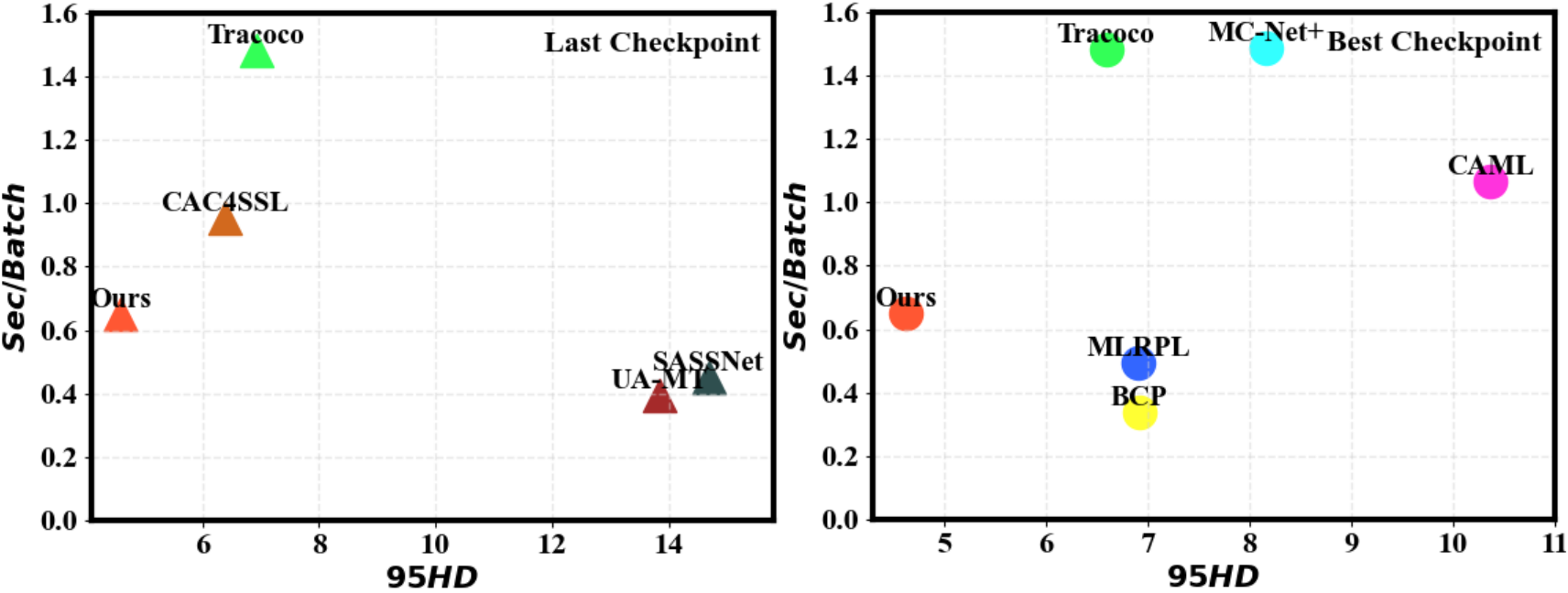}
\caption{Comparative results of training times under the 10\% partition protocol on the LA dataset.}
\label{fig:training_time}
\end{figure}

\textbf{Training Comparison.} We compare the training time of our method with other SOTA methods on the LA dataset under the 10\% partition protocol in Fig.~\ref{fig:training_time}. All the experiments are implemented on the Ubuntu system with NVIDIA GeForce RTX 3090 GPU, where the batch size is set to 4, and all volumes are cropped to 112 $\times$ 112 $\times$ 80. As illustrated in Fig.~\ref{fig:training_time}, the training time of our method on a single GPU is 0.651 sec/batch, a significant improvement over other state-of-the-art methods such as CAC4SSL~\citep{li2024contour}, TraCoCo~\citep{liu2022translation} and  CAML~\citep{gao2023correlation}. Compared with the teacher-student-based models like BCP~\citep{bai2023bidirectional} and UA-MT~\citep{yu2019uncertainty}, our approach only marginally increases the training time, while significantly enhancing segmentation performance.


\begin{figure}[!t]
\centering
\includegraphics[width=3.5in]{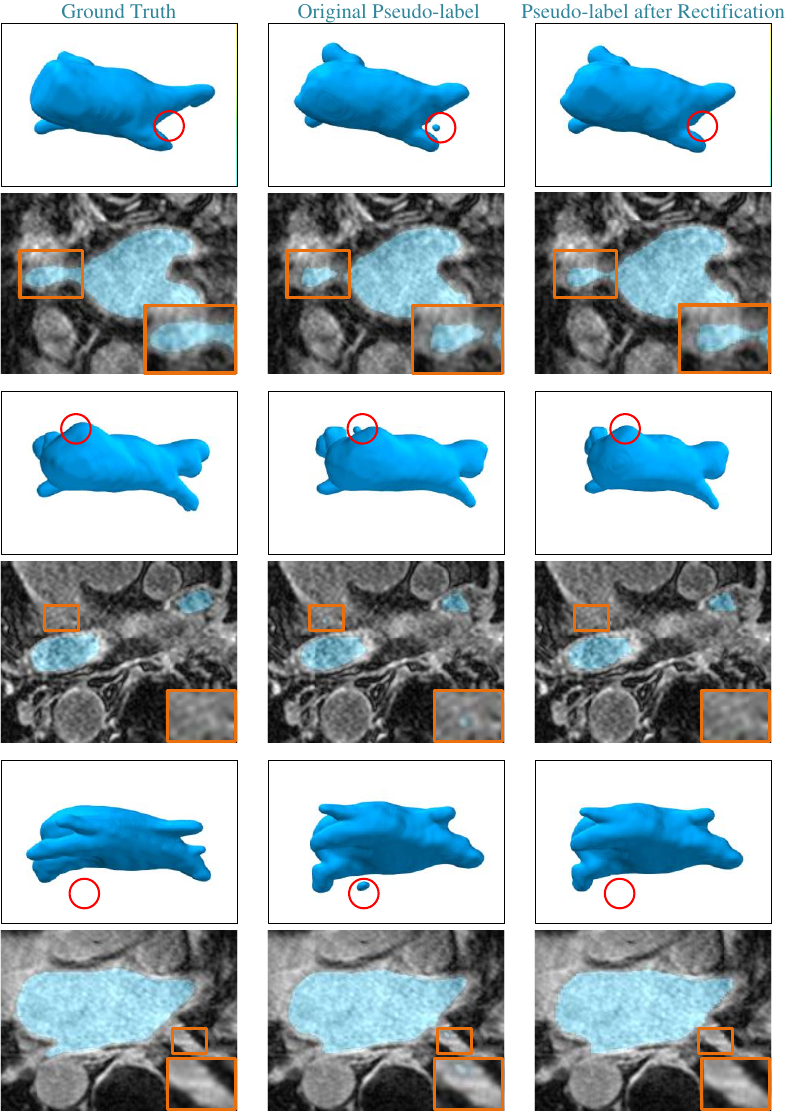}
\caption{Visualisations of rectified pseudo-labels using the CRLN equipped with  DIM at the last training iteration. We highlight the mistakes corrected by the rectification process in the segmentation images and volumes.}
\label{fig:Visualisations_corrected_pseudo-labels}
\end{figure}

\subsection{Ablation Study of Key Components}\label{sec:Ablation_Study}

The roles of our crucial components are evaluated and analysed on LA and Pancreas-CT datasets under the 10\% partition protocol. 
Tab.~\ref{tab:table5} illustrates the gains brought by each key module compared to the baseline, where 
the teacher-student framework equipped with weak-to-strong consistency regularisation serves as our baseline.
\textcolor{red}{We first study the roles of strong augmentation (CutMix+Random Noise), the Cooperative Rectification Learning Network (CRLN) equipped with the Dynamic Interaction Module (DIM), and Collaborative Positive Supervision (CPS). Subsequently, we investigate the role of DIM alone and evaluate the impact of key parameters.}
\textcolor{red}{\subsubsection{Ablation Study of the Strong Augmentation}}\label{sec:strong_augmentation}
\textcolor{red}{An appropriately strong augmentation strategy is crucial for effectively leveraging consistency constraints in training the segmentation model. Instead of relying solely on a single strong augmentation technique, such as centre cropping used in the baseline, we employ multiple strong augmentation methods, including centre cropping, CutMix, and random noise. As shown in Table~\ref{tab:table5}, CutMix and random noise provide considerable improvements compared to the baseline. For example, the Dice is improved by 4.48\% on the LA dataset. These enhancements demonstrate the advantage of diverse augmentation strategies in boosting the segmentation performance by introducing varied perturbations.}
\textcolor{red}{\subsubsection{Ablation Study of the CRLN equipped with DIM}}\label{sec:CRLN_DIM}
The CRLN equipped with the DIM focuses on leveraging the label data knowledge to rectify pseudo-labels, thus allowing the use of valuable unlabelled information to influence training with an accurate supervision for the student model. \textcolor{red}{As evidenced in Tab.~\ref{tab:table5}, the CRLN integrated with DIM improves the model by a clear margin. The 95HD is improved by 4.33 on the LA dataset, and the Dice is raised by 1.85\% on the Pancreas-CT dataset.} Such improvement demonstrates the effectiveness of our proposed CRLN+DIM. 

\textcolor{red}{To gain deeper insight into CRLN+DIM, we investigate the role of correcting pseudo-labels during training. Fig.~\ref{fig:quality_pseudo_label} presents a quantitative comparison between the original pseudo-labels and the pseudo-labels after rectification in the early training stages. As shown in Fig.~\ref{fig:quality_pseudo_label}(a), there is a significant increase in reliable predictions in the corrected pseudo-labels. For instance, the percentage of reliable predictions rises by 22.7\% at the 1400$^{th}$ iteration. These improvements indicate that many initially unreliable predictions have been rectified, allowing more unlabelled information to contribute actively to the training process. Additionally, the prediction accuracy achieved using the rectified pseudo-labels is substantially better than that of the original (and reliable) pseudo-labels, as illustrated in Fig.~\ref{fig:quality_pseudo_label}(b). The Dice score of the reliable pseudo-labels increases from 88.51\% to 91.32\% at the 1800$^{th}$ training iteration. This demonstrates that CRLN+DIM indeed provides more accurate supervision signals for training on unlabelled data, thereby enhancing semi-supervised segmentation performance.}

\begin{figure}[!t]
\centering
\includegraphics[width=3.5in]{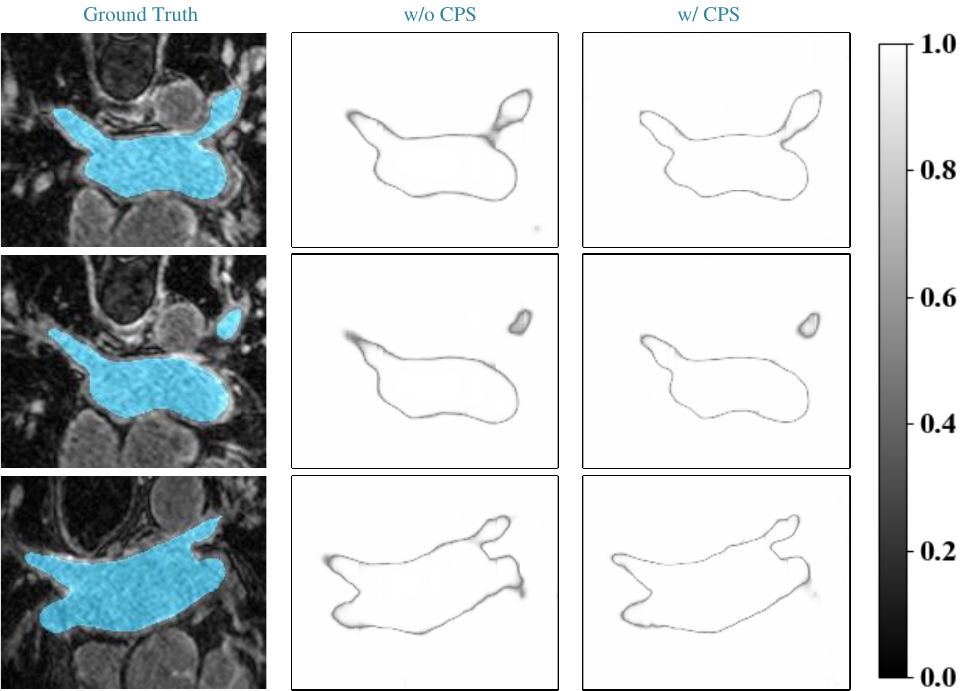}
\caption{\textcolor{red}{Comparison of confidence maps with and without the CPS mechanism.}}
\label{fig:Visualisations_boundary}
\end{figure}

Furthermore, Fig.~\ref{fig:Visualisations_corrected_pseudo-labels} visualises both the original and rectified pseudo-labels of the unlabelled images from the LA dataset at the last training iteration. It can be seen that the problem of incorrectly segmenting the background into the left atrium is present in the original pseudo-labels. Through learnable pairwise interactions, the CRLN+DIM can leverage the labelled information to correct the original teacher's predictions. As shown in the third column of Fig.~\ref{fig:Visualisations_corrected_pseudo-labels}, the CRLN+DIM can suppress the background interference and effectively rectify the incorrect prediction in the original pseudo-labels.
\textcolor{red}{\subsubsection{Ablation Study of the CPS}}\label{sec:CPS}
The proposed CPS mechanism aims to enhance the representation of hard regions using unassertive positive samples, as explained in~\eqref{eq:collaborative_center}. As shown in Tab.~\ref{tab:table5}, consistent improvements are achieved by the CPS mechanism on both the LA and Pancreas-CT datasets. \textcolor{red}{For instance, the Dice value is increased by 1.87\% and 0.59\%, and the 95HD is decreased by 2.04 and 2.56 on the two datasets, respectively. Moreover, we visualize the confidence maps of the model's predictions in Fig.~\ref{fig:Visualisations_boundary}. It can be seen that the CPS mechanism indeed improves the model's prediction quality in difficult regions, such as boundary regions.}

\textcolor{red}{To better understand the impact of the CPS mechanism, we examine how unassertive positive samples—created by combining class mean representations with learned prototypes—contribute to its performance. This investigation compares our approach with two alternative strategies: one utilising solely class mean representations for the positive samples, while the other exclusively employing the learned prototypes ($\mathsf{mean}(P^c)$ in ~\eqref{eq:collaborative_center}).}
Tab.~\ref{tab:table5} reports the results. It can be observed that our strategy consistently outperforms the other two strategies, demonstrating the benefits of our moderate strategy.  


\begin{table}[!t]
\caption{The performance of different aggregation mechanisms in the DIM under the 10\% partition protocol on the LA dataset. "SA" denotes the spatial awareness, "CI" is the convolutional integration, and "CR" represents the cross-class reasoning.\label{tab:table6}}
\centering
\setlength{\tabcolsep}{0.73mm}{
\begin{tabular}{c c c| c c c c }
\toprule[1pt]
\textcolor{red}{SA} & \textcolor{red}{CI} & \textcolor{red}{CR} & \textcolor{red}{Dice(\%)} & \textcolor{red}{Jaccard(\%)} & \textcolor{red}{ASD(Voxel)} & \textcolor{red}{95HD(Voxel)} \\
\hline
 & & & \textcolor{red}{90.22\scriptsize$\pm$0.90} & \textcolor{red}{82.42\scriptsize$\pm$1.33} & \textcolor{red}{1.58\scriptsize$\pm$0.11} & \textcolor{red}{6.06\scriptsize$\pm$0.77} \\
\textcolor{red}{\checkmark} & & & \textcolor{red}{90.89\scriptsize$\pm$0.54} & \textcolor{red}{83.40\scriptsize$\pm$0.88} & \textcolor{red}{1.55\scriptsize$\pm$0.12} & \textcolor{red}{5.44\scriptsize$\pm$0.43} \\
\textcolor{red}{\checkmark} & \textcolor{red}{\checkmark} & & \textcolor{red}{91.04\scriptsize$\pm$0.40} & \textcolor{red}{83.61\scriptsize$\pm$0.67} & \textcolor{red}{1.46\scriptsize$\pm$0.09} & \textcolor{red}{5.26\scriptsize$\pm$0.38} \\
\textcolor{red}{\checkmark} & \textcolor{red}{\checkmark} & \textcolor{red}{\checkmark} & \textcolor{red}{\textbf{91.74}\scriptsize$\pm$0.38} & \textcolor{red}{\textbf{84.79}\scriptsize$\pm$0.64} & \textcolor{red}{\textbf{1.46}\scriptsize$\pm$0.09} & \textcolor{red}{\textbf{4.60}\scriptsize$\pm$0.33} \\
\bottomrule[1pt]
\end{tabular}}
\end{table}

\textcolor{red}{\subsubsection{Ablation Study of the Aggregation Strategy at DIM}}

The spatial-aware and cross-class aggregation strategy in~\eqref{eq:spatial_aware_cross_class_aggregation} is proposed in the DIM module to provide accurate clues for correction. The proposed aggregation strategy contains three unique designs: Spatial Awareness (SA, denoted by the $\mathsf{Conv}_{3 \times 3 \times 3}(\cdot)$ in~\eqref{eq:spatial_aware_cross_class_aggregation}), Convolutional Integration (CI, represented by the $\mathsf{Conv}_{1 \times 1 \times 1}(\cdot)$ in~\eqref{eq:spatial_aware_cross_class_aggregation}), and Cross-class Reasoning (CR, denoted by the sharing of convolutional parameters between different categories).
\textcolor{red}{As shown in Tab.~\ref{tab:table6}, SA demonstrates a clear improvement over the baseline based on the direct summation of the proximity matrices, with Jaccard increasing by 0.98\% and 95HD decreasing by 0.62.} Such improvements show the benefits of re-evaluating the relationships between features and each prototype group based on the local context. \textcolor{red}{Additionally, CI yields a slight boost on all metrics. The dependency reasoning across different classes (i.e., CR) further raises Dice to 91.74\% and decreases 95HD to 4.60, demonstrating the effectiveness of the information interaction across categories.}

\begin{figure}[!t]
\centering
\includegraphics[width=3.5in]{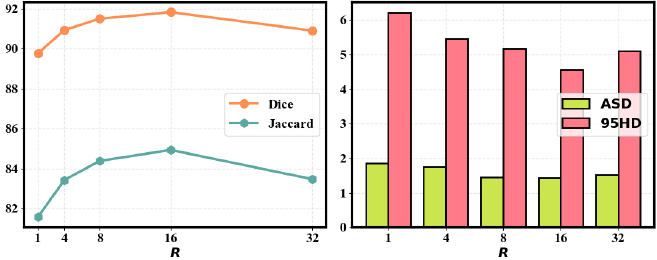}
\caption{The segmentation quality of the method as a function of the number $R$ of prototypes for the 10\% partition protocol on the LA dataset.}
\label{fig:different_numbers_ of_prototypes}
\end{figure}

\textcolor{red}{\subsubsection{Ablation Study of the Necessity of Multiple Prototypes for Each Category}}

As discussed in Sec.~\ref{sec:pdr}, we argue that it is essential to construct multiple prototypes per class. To validate this point, the segmentation quality of the method under different numbers of prototypes is analysed. As illustrated in Fig.~\ref{fig:different_numbers_ of_prototypes}, representing a class with multiple prototypes achieves superior results compared to using only one prototype. \textcolor{red}{For instance, the Jaccard with multiple prototypes improves by approximately 1.86\% to 3.37\% compared to a single prototype, demonstrating that one prototype is insufficient to characterise complex medical data.} \textcolor{red}{In addition, the results in Fig.~\ref{fig:different_numbers_ of_prototypes} also indicate that increasing the number of prototypes does not always lead to better performance. When the number of prototypes is increased to 32, the model's performance declines compared to using 16 prototypes, with a decrease of 1.47\% in Jaccard and an increase of 0.55 in 95HD.} This is mainly because the proper number of prototypes can extract discriminative features of the class, but too many prototypes can cause the model to overfit the labelled data distribution.

\textcolor{red}{\subsubsection{Ablation Study of Different Rectification Mechanisms at CRLN}}

The goal of CRLN's rectification mechanism in~\eqref{eq:rectification_student} is to correct the original prediction under the guidance of the prototypes learned from the labelled data. 
Three different versions are designed to find a suitable rectification mechanism. 
The r-fixed-v1 replaces the low-confidence original pseudo-labels of the unlabelled samples with the holistic relationship map $\mathscr{M}(x)$ if the prediction confidence of $\mathscr{M}(x)$ exceeds that of the original pseudo-labels $\bar{y}$, as follows:
\begin{equation}
  \bar{y}_{r} = \begin{cases}
    \mathscr{M}(x) & , \text{if $(\max \bar{y} < \tau)$ and $(\max \bar{y} < \max \mathscr{M}(x))$}\\
    \bar{y} & , \text{otherwise}.
  \end{cases}
\end{equation}
The other two strategies modify the original pseudo-labels in a learnable manner.
\textcolor{red}{The r-learnable-v2 concatenates the $\bar{y}$ and $\mathscr{M}(x)$, followed by a $3 \times 3 \times 3$ convolution layer.}
The r-learnable-v3 consists of our proposed approach, explained in Sec.~\ref{sec:rectification_stage}, where the rectification corrects the predicted label with $\bar{y}_{r} = \bar{y} + (1-\mu)\times \mathscr{M}(x)$. 
The results of the whole network with different rectification mechanisms
are shown in Tab.~\ref{tab:table8}. 
It is clear that the r-learnable-v3 demonstrates superior rectification compared to the other strategies. Therefore, the r-learnable-v3 is selected as the final rectification strategy in our paper.

\begin{figure}[!t]
\centering
\includegraphics[width=3.5in]{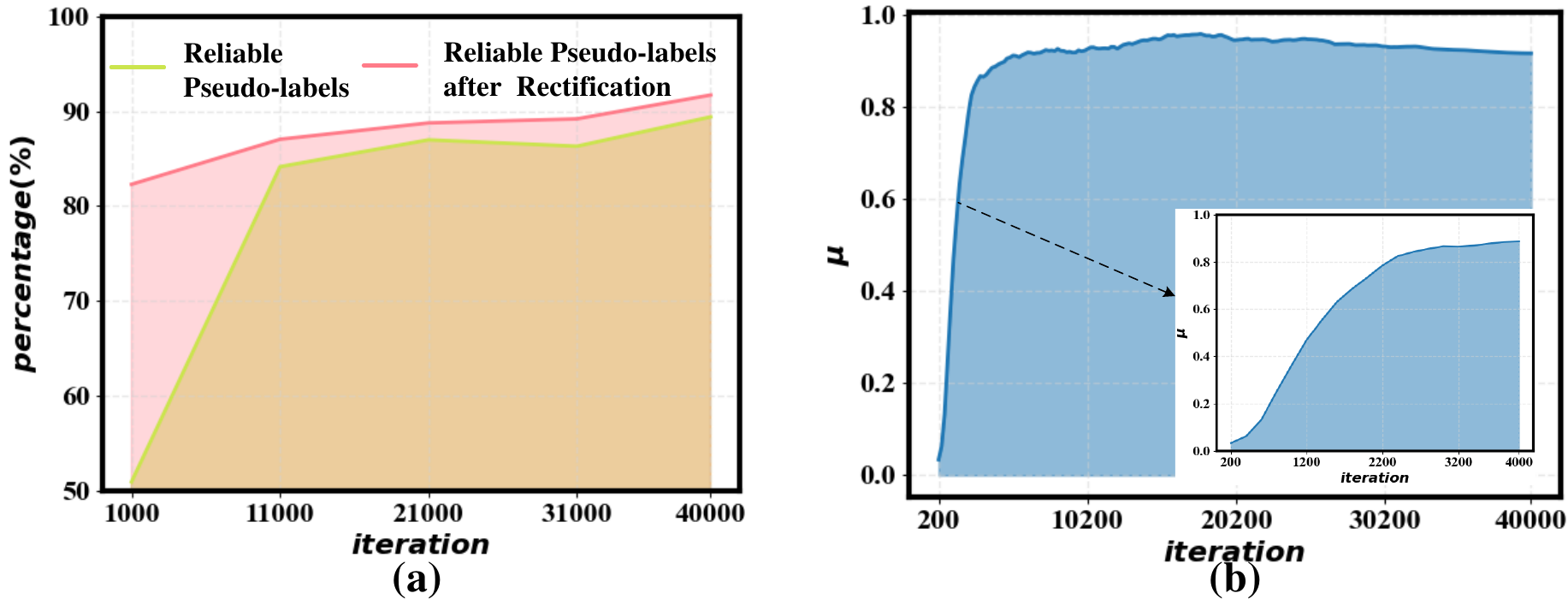}
{\caption{\textcolor{red}{The quality of pseudo-labels and the value of $\mu$ as functions of training iterations under the 10\% partition protocol on the LA dataset.}}
\label{fig:mu_train}}
\end{figure}

\textcolor{red}{The learnable parameter $\mu$ in our rectification strategy is utilised to regulate the extent of the correction. As shown in Fig. ~\ref{fig:mu_train}, during the early training stages, the pseudo-labels are highly unreliable, so the value of $\mu$ is low, indicating a strong correction guided by labelled knowledge priors. As the model continues to learn, the quality of the pseudo-labels improves, leading to a reduced need for correction, which is reflected in the increasing value of $\mu$.}



\begin{table}[!t]
\caption{Performance of CRLN with different rectification mechanisms under the 10\% partition protocol on the LA dataset.\label{tab:table8}}
\centering
\setlength{\tabcolsep}{0.35mm}{
\begin{tabular}{ c c c c c }
\toprule[1pt]
\textcolor{red}{Method}&\textcolor{red}{Dice(\%)}&\textcolor{red}{Jaccard(\%)}&\textcolor{red}{ASD(Voxel)}&\textcolor{red}{95HD(Voxel)}\\
\hline
\textcolor{red}{r-fixed-v1}& \textcolor{red}{91.14\scriptsize$\pm$0.43} & \textcolor{red}{83.79\scriptsize$\pm$0.71} & \textcolor{red}{1.56\scriptsize$\pm$0.09} & \textcolor{red}{5.07\scriptsize$\pm$0.38}\\
\textcolor{red}{r-learnable-v2} & \textcolor{red}{90.69\scriptsize$\pm$0.64} & \textcolor{red}{83.09\scriptsize$\pm$1.01} & \textcolor{red}{1.66\scriptsize$\pm$0.13} & \textcolor{red}{5.58\scriptsize$\pm$0.45}\\
\textcolor{red}{r-learnable-v3}& \textcolor{red}{\textbf{91.74}\scriptsize$\pm$0.38} & \textcolor{red}{\textbf{84.79}\scriptsize$\pm$0.64} & \textcolor{red}{\textbf{1.46}\scriptsize$\pm$0.09} & \textcolor{red}{\textbf{4.60}\scriptsize$\pm$0.33}\\
\bottomrule[1pt]
\end{tabular}}
\end{table}

\begin{figure}[!t]
\centering
\includegraphics[width=3.5in]{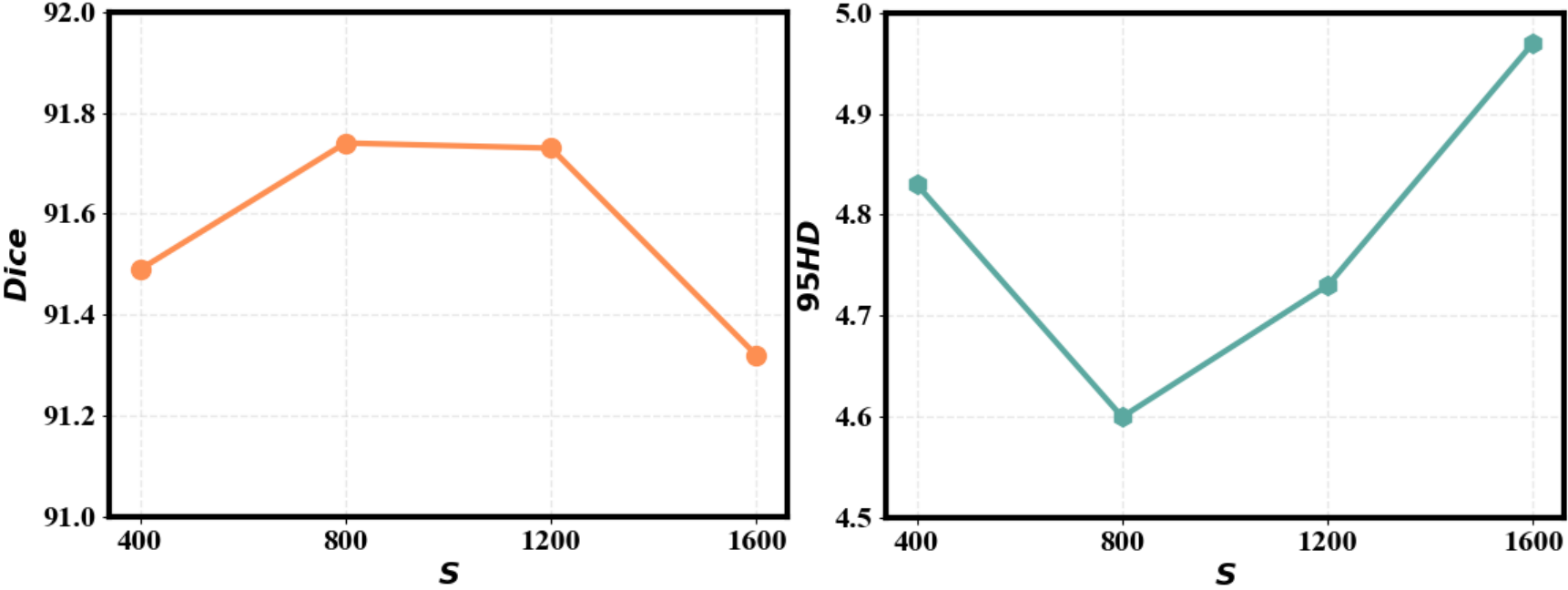}
{\caption{The segmentation performance of our method with different values of $S$ under the 10\% partition protocol on the LA dataset. 
}
\label{fig:different_values_ of_s}}
\end{figure}

\begin{figure}[!t]
\centering
\includegraphics[width=3.5in]{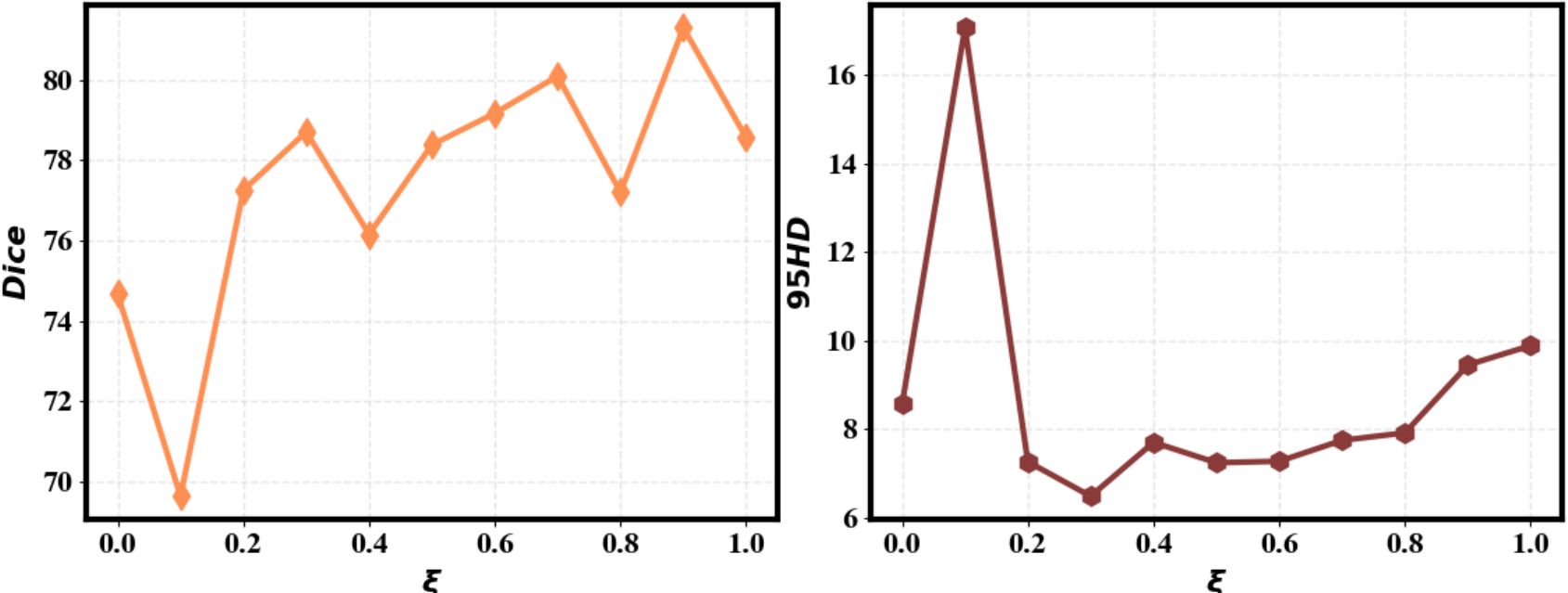}
{\caption{The segmentation performance of our method with different values of coefficient $\xi$ under 10\% partition protocol on the  Pancreas-CT dataset. 
}
\label{fig:different_values_ of_coefficient}}
\end{figure}
\begin{figure*}[!t]
\centering
\includegraphics[width=6.5in]{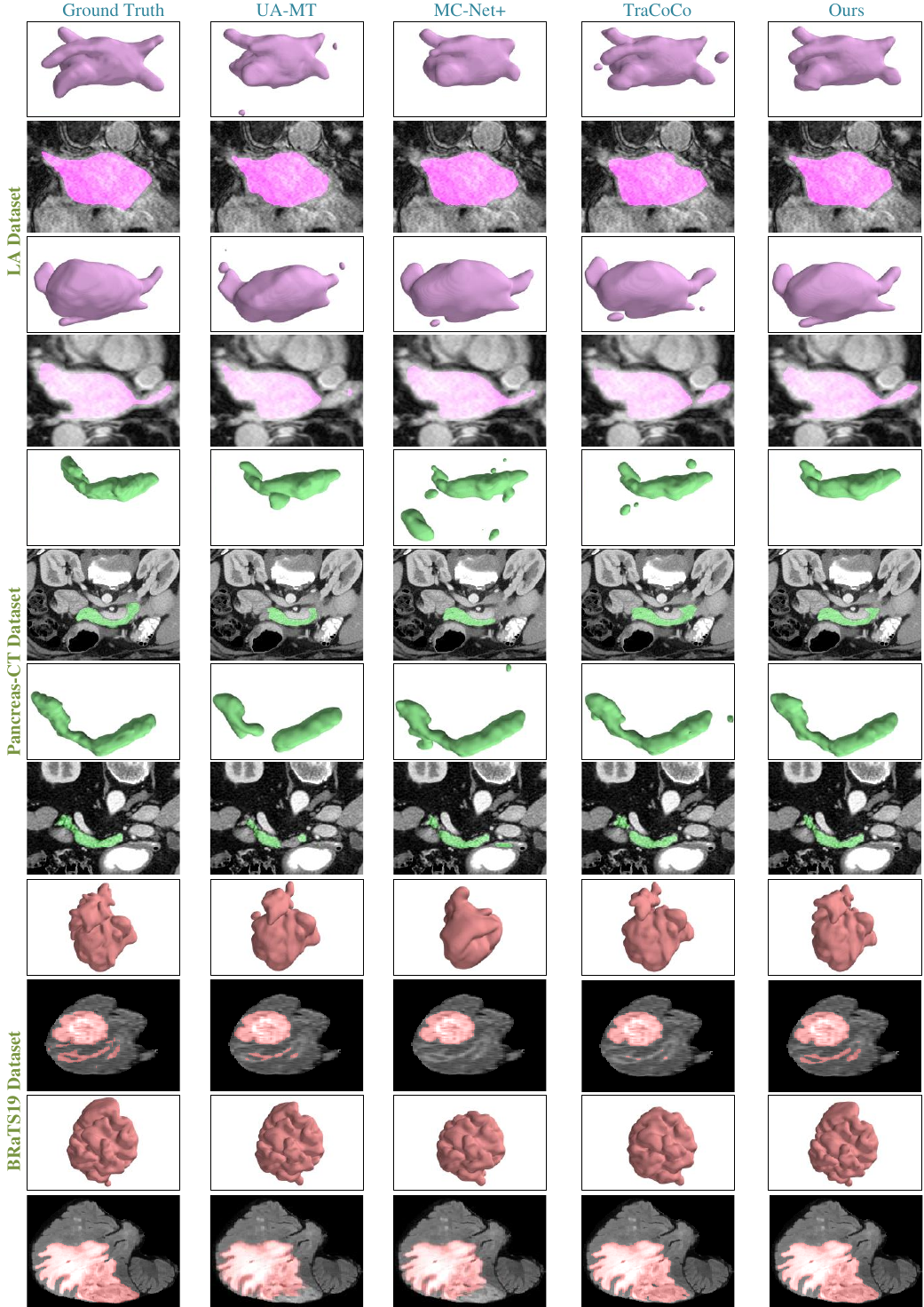}
{\caption{{Visualisation of the results from different methods for the 10\% partition protocol on the LA, Pancreas-CT, and BraTS19s datasets.}}
\label{fig:visualised_results}}
\end{figure*}

\textcolor{red}{\subsubsection{Ablation Study of the Hyper-parameter $S$ at CRLN}}
\label{sec:ablation_s}
As discussed in Sec.~\ref{sec:rectification_stage}, the hyper-parameter $S$ determines when to start correcting pseudo-labels using the learned category prototypes. We examine different values of $S$ from 
$\{400,800,1200,1600\}$
to observe the whole model performance. As shown in Fig.~\ref{fig:different_values_ of_s}, the model maintains consistent performance across various values of $S$, especially at $S=800$, where it works best.
\textcolor{red}{\subsubsection{Ablation Study of the Hyper-parameter $\xi$ in ~\eqref{eq:collaborative_center}}
\label{sec:ablation_xi}}
The hyper-parameter $\xi$ in ~\eqref{eq:collaborative_center} controls the trade-off between the prototypes and mean representations, which is crucial in the collaborative positive supervision mechanism. Consequently, we analyse the effect of different values of $\xi$ on the segmentation performance of the overall model. Experimental results are illustrated in Fig.~\ref{fig:different_values_ of_coefficient}. It can be observed that the proposed method exhibits competitive results when $\xi$ ranges from 0.2 to 0.8. Considering these results, $\xi$ is set to 0.6.

\textcolor{red}{\subsection{Comparison with the SAM Models}}
\textcolor{red}{Large models, such as SAM (Segment Anything Model), are gaining popularity in medical image analysis due to their strong generalization capabilities. We compare the proposed method with state-of-the-art SAM models under the same annotation rate. As shown in Tab.~\ref{tab:com_sam_la_10}, the proposed method consistently achieves superior performance. For example, compared to Auto-SAM \citep{shaharabany2023autosam} and MA-SAM \citep{chen2024ma}, the 95HD is decreased by 26.94 and 1.75, respectively. These results demonstrate the effectiveness of the proposed method.}
\subsection{Visualisation Results}
\begin{figure}[!t]
\centering
\includegraphics[width=3.5in]{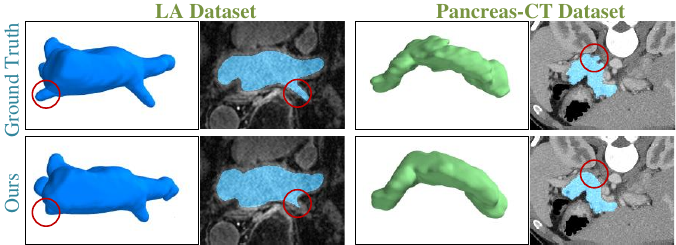}
{\caption{~\textcolor{red}{Visualisation of the failure cases from our method for the 10\% partition protocol on the LA and Pancreas-CT datasets. We highlight the regions that are poorly segmented by our method in images and volumes.}}
\label{fig:failure_case}}
\end{figure}

Fig.~\ref{fig:visualised_results} displays the 3D and 2D visualisations of the proposed method, UA-MT \citep{yu2019uncertainty}, MC-Net+\citep{wu2022mutual} and TraCoCo \citep{liu2022translation} on the LA, Pancreas-CT and BRaTS2019 datasets. The partition protocol is 10\% on all three datasets. Benefiting from the leveraging of the labelled data, the segmentation results of the proposed method are closer to the ground truth than other methods. As shown in the 3D visualisation results, particularly in the third and fifth rows of Fig.~\ref{fig:visualised_results}, the proposed method results in a more complete edge segmentation while suppressing the interferences in the backgrounds. Furthermore, from the tenth row of Fig.~\ref{fig:visualised_results}, it can be observed that our method maintains segmentation integrity even when dealing with tiny tissues with complex shapes. Such visual differences demonstrate that the proposed method is effective in improving the segmentation of challenging regions. 
\textcolor{red}{\subsection{Limitations and Future Work}}
\textcolor{red}{While the experimental results confirm the effectiveness of our method, there are still some limitations that are worth discussing. 1) The segmentation effectiveness of our method declines in fine regions of certain anatomies, such as in neighbouring parts of the pulmonary veins shown in Fig.\ref{fig:failure_case}. This issue may arise from the small size and low pixel count of these regions, which provide insufficient information for the model to effectively differentiate them from surrounding tissues. 2) Our method can produce segmentation results that may be over-smoothed in some regions, which may not accurately capture the natural folds and curvature variations of the pancreatic surface, as depicted in the third and fourth columns of Fig.\ref{fig:failure_case}. This limitation could be due to the model's insensitivity to high-frequency variations, with a tendency to focus on low-frequency global features during training. 3) Our current work primarily focuses on evaluating the proposed semi-supervised segmentation method in medical segmentation tasks and does not directly assess its impact on downstream clinical applications.}

~\textcolor{red}{Based on these limitations, future work will focus on increasing the model's sensitivity to fine details by designing shape- or detail-aware loss functions. Additionally, developing specialized data augmentation techniques for small regions and complex surfaces to expand the diversity and representation of such samples in the training dataset will be an interesting research topic. Furthermore, more precise segmentation could provide more accurate tumour features—such as volume, shape, and texture—which are crucial for predicting patient outcomes. To further assess the practical impact of our improvements on clinical prediction and decision-making, we plan to extend our approach to downstream clinical tasks, such as survival analysis.}
\begin{table}[!t]
\caption{\textcolor{red}{Comparative results with SAM models under the 10\% partition protocol on the LA dataset.}\label{tab:com_sam_la_10}}
\centering
\setlength{\tabcolsep}{0.5mm}{
\begin{tabular}{ c c c c c }
\toprule[1pt]
\textcolor{red}{Method}&\textcolor{red}{Dice(\%)}&\textcolor{red}{Jaccard(\%)}&\textcolor{red}{ASD(Voxel)}&\textcolor{red}{95HD(Voxel)}\\
\hline
\textcolor{red}{AutoSAM} &\textcolor{red}{48.94\scriptsize{$\pm$2.85}} & \textcolor{red}{33.31\scriptsize{$\pm$2.38}} &\textcolor{red}{11.90\scriptsize{$\pm$0.56}} & \textcolor{red}{31.54\scriptsize{$\pm$1.56}}\\
\textcolor{red}{MA-SAM} &\textcolor{red}{89.66\scriptsize{$\pm$0.77}}& \textcolor{red}{81.43\scriptsize{$\pm$1.21}} & \textcolor{red}{2.15\scriptsize{$\pm$0.16}} & \textcolor{red}{6.35\scriptsize{$\pm$0.63}}\\
\textcolor{red}{Ours}& \textcolor{red}{{\textbf{91.74}}\scriptsize{$\pm$0.38}}& \textcolor{red}{{\textbf{84.79}}\scriptsize{$\pm$0.64}}&\textcolor{red}{{\textbf{1.46}}\scriptsize{$\pm$0.09}}&\textcolor{red}{{\textbf{4.60}}\scriptsize{$\pm$0.33}}\\
\bottomrule[1pt]
\end{tabular}}
\end{table}

\section{Conclusion}\label{sec:Conclusion}

In this paper, we proposed a new semi-supervised 3D medical image segmentation method, which leverages the labelled information as external priors to improve the prediction quality of pseudo-labels and boosts the segmentation accuracy of uncertain regions. Specifically, the cooperative rectification learning network based on the dynamic interaction module first captures multiple prototypical priors for each class from the labelled data and then utilises these priors to improve the segmentation accuracy of pseudo-labels in a learnable pairwise interaction pattern. Such a design provides more precise supervision for the student model, enabling the use of a large amount of unlabelled data in training. Moreover, the collaborative positive supervision mechanism encourages the uncertain samples to move closer to the unassertive positive learning samples, thereby endowing the model with the ability to distinguish uncertain regions. Compared to SOTA semi-supervised 3D medical image segmentation approaches, our proposed network achieves more accurate segmentation performance on LA, Pancreas and BRaTS19 datasets. 

\textcolor{red}{Despite these promising results, the proposed method still has limitations, particularly in segmenting detailed parts of certain anatomies, and lacks a direct assessment of its impact on downstream clinical applications. Future work will focus on improving segmentation performance for detailed parts of particular tissues by designing shape- or detail-aware loss functions and developing specialised data augmentation techniques for small regions and complex surfaces. Additionally, we plan to extend our approach to downstream clinical tasks, such as survival analysis, to better evaluate the practical impact of our improvements on clinical prediction and decision-making.}


\section*{Acknowledgements}
This work is supported by the China Scholarship Council (202306080128).
\bibliographystyle{model2-names.bst}\biboptions{authoryear}
\bibliography{refs}

\begin{thebibliography}{55}
\expandafter\ifx\csname natexlab\endcsname\relax\def\natexlab#1{#1}\fi
\providecommand{\url}[1]{\texttt{#1}}
\providecommand{\href}[2]{#2}
\providecommand{\path}[1]{#1}
\providecommand{\DOIprefix}{doi:}
\providecommand{\ArXivprefix}{arXiv:}
\providecommand{\URLprefix}{URL: }
\providecommand{\Pubmedprefix}{pmid:}
\providecommand{\doi}[1]{\href{http://dx.doi.org/#1}{\path{#1}}}
\providecommand{\Pubmed}[1]{\href{pmid:#1}{\path{#1}}}
\providecommand{\bibinfo}[2]{#2}
\ifx\xfnm\relax \def\xfnm[#1]{\unskip,\space#1}\fi
\bibitem[{{Adiga V.} et~al.(2024){Adiga V.}, Dolz and Lombaert}]{ADIGAV2024103011}
\bibinfo{author}{{Adiga V.}, S.}, \bibinfo{author}{Dolz, J.}, \bibinfo{author}{Lombaert, H.}, \bibinfo{year}{2024}.
\newblock \bibinfo{title}{Anatomically-aware uncertainty for semi-supervised image segmentation}.
\newblock \bibinfo{journal}{Medical Image Analysis} \bibinfo{volume}{91}, \bibinfo{pages}{103011}.
\newblock \URLprefix \url{https://www.sciencedirect.com/science/article/pii/S1361841523002712}, \DOIprefix\doi{https://doi.org/10.1016/j.media.2023.103011}.
\bibitem[{Bai et~al.(2023)Bai, Chen, Li, Shen and Wang}]{bai2023bidirectional}
\bibinfo{author}{Bai, Y.}, \bibinfo{author}{Chen, D.}, \bibinfo{author}{Li, Q.}, \bibinfo{author}{Shen, W.}, \bibinfo{author}{Wang, Y.}, \bibinfo{year}{2023}.
\newblock \bibinfo{title}{Bidirectional copy-paste for semi-supervised medical image segmentation}, in: \bibinfo{booktitle}{Proceedings of the IEEE/CVF Conference on Computer Vision and Pattern Recognition}, pp. \bibinfo{pages}{11514--11524}.
\bibitem[{Chen et~al.(2024)Chen, Miao, Wu, Zhong, Yan, Kim, Hu, Liu, Sun, Li et~al.}]{chen2024ma}
\bibinfo{author}{Chen, C.}, \bibinfo{author}{Miao, J.}, \bibinfo{author}{Wu, D.}, \bibinfo{author}{Zhong, A.}, \bibinfo{author}{Yan, Z.}, \bibinfo{author}{Kim, S.}, \bibinfo{author}{Hu, J.}, \bibinfo{author}{Liu, Z.}, \bibinfo{author}{Sun, L.}, \bibinfo{author}{Li, X.}, et~al., \bibinfo{year}{2024}.
\newblock \bibinfo{title}{Ma-sam: Modality-agnostic sam adaptation for 3d medical image segmentation}.
\newblock \bibinfo{journal}{Medical Image Analysis} , \bibinfo{pages}{103310}.
\bibitem[{Chen et~al.(2021)Chen, Lu, Yu, Luo, Adeli, Wang, Lu, Yuille and Zhou}]{chen2021transunet}
\bibinfo{author}{Chen, J.}, \bibinfo{author}{Lu, Y.}, \bibinfo{author}{Yu, Q.}, \bibinfo{author}{Luo, X.}, \bibinfo{author}{Adeli, E.}, \bibinfo{author}{Wang, Y.}, \bibinfo{author}{Lu, L.}, \bibinfo{author}{Yuille, A.L.}, \bibinfo{author}{Zhou, Y.}, \bibinfo{year}{2021}.
\newblock \bibinfo{title}{Transunet: Transformers make strong encoders for medical image segmentation}.
\newblock \bibinfo{journal}{arXiv preprint arXiv:2102.04306} .
\bibitem[{Chen et~al.(2022)Chen, Zhang, Debattista and Han}]{chen2022semi}
\bibinfo{author}{Chen, J.}, \bibinfo{author}{Zhang, J.}, \bibinfo{author}{Debattista, K.}, \bibinfo{author}{Han, J.}, \bibinfo{year}{2022}.
\newblock \bibinfo{title}{Semi-supervised unpaired medical image segmentation through task-affinity consistency}.
\newblock \bibinfo{journal}{IEEE Transactions on Medical Imaging} \bibinfo{volume}{42}, \bibinfo{pages}{594--605}.
\bibitem[{Chen et~al.(2019)Chen, Bortsova, Garc{\'\i}a-Uceda~Ju{\'a}rez, Van~Tulder and De~Bruijne}]{chen2019multi}
\bibinfo{author}{Chen, S.}, \bibinfo{author}{Bortsova, G.}, \bibinfo{author}{Garc{\'\i}a-Uceda~Ju{\'a}rez, A.}, \bibinfo{author}{Van~Tulder, G.}, \bibinfo{author}{De~Bruijne, M.}, \bibinfo{year}{2019}.
\newblock \bibinfo{title}{Multi-task attention-based semi-supervised learning for medical image segmentation}, in: \bibinfo{booktitle}{Medical Image Computing and Computer Assisted Intervention--MICCAI 2019: 22nd International Conference, Shenzhen, China, October 13--17, 2019, Proceedings, Part III 22}, \bibinfo{organization}{Springer}. pp. \bibinfo{pages}{457--465}.
\bibitem[{Chowdary and Yin(2023)}]{chowdary2023diffusion}
\bibinfo{author}{Chowdary, G.J.}, \bibinfo{author}{Yin, Z.}, \bibinfo{year}{2023}.
\newblock \bibinfo{title}{Diffusion transformer u-net for medical image segmentation}, in: \bibinfo{booktitle}{International Conference on Medical Image Computing and Computer-Assisted Intervention}, \bibinfo{organization}{Springer}. pp. \bibinfo{pages}{622--631}.
\bibitem[{{\c{C}}i{\c{c}}ek et~al.(2016){\c{C}}i{\c{c}}ek, Abdulkadir, Lienkamp, Brox and Ronneberger}]{cciccek20163d}
\bibinfo{author}{{\c{C}}i{\c{c}}ek, {\"O}.}, \bibinfo{author}{Abdulkadir, A.}, \bibinfo{author}{Lienkamp, S.S.}, \bibinfo{author}{Brox, T.}, \bibinfo{author}{Ronneberger, O.}, \bibinfo{year}{2016}.
\newblock \bibinfo{title}{3d u-net: learning dense volumetric segmentation from sparse annotation}, in: \bibinfo{booktitle}{Medical Image Computing and Computer-Assisted Intervention--MICCAI 2016: 19th International Conference, Athens, Greece, October 17-21, 2016, Proceedings, Part II 19}, \bibinfo{organization}{Springer}. pp. \bibinfo{pages}{424--432}.
\bibitem[{Clark et~al.(2013)Clark, Vendt, Smith, Freymann, Kirby, Koppel, Moore, Phillips, Maffitt, Pringle et~al.}]{clark2013cancer}
\bibinfo{author}{Clark, K.}, \bibinfo{author}{Vendt, B.}, \bibinfo{author}{Smith, K.}, \bibinfo{author}{Freymann, J.}, \bibinfo{author}{Kirby, J.}, \bibinfo{author}{Koppel, P.}, \bibinfo{author}{Moore, S.}, \bibinfo{author}{Phillips, S.}, \bibinfo{author}{Maffitt, D.}, \bibinfo{author}{Pringle, M.}, et~al., \bibinfo{year}{2013}.
\newblock \bibinfo{title}{The cancer imaging archive (tcia): maintaining and operating a public information repository}.
\newblock \bibinfo{journal}{Journal of digital imaging} \bibinfo{volume}{26}, \bibinfo{pages}{1045--1057}.
\bibitem[{Dosovitskiy et~al.(2020)Dosovitskiy, Beyer, Kolesnikov, Weissenborn, Zhai, Unterthiner, Dehghani, Minderer, Heigold, Gelly et~al.}]{dosovitskiy2020image}
\bibinfo{author}{Dosovitskiy, A.}, \bibinfo{author}{Beyer, L.}, \bibinfo{author}{Kolesnikov, A.}, \bibinfo{author}{Weissenborn, D.}, \bibinfo{author}{Zhai, X.}, \bibinfo{author}{Unterthiner, T.}, \bibinfo{author}{Dehghani, M.}, \bibinfo{author}{Minderer, M.}, \bibinfo{author}{Heigold, G.}, \bibinfo{author}{Gelly, S.}, et~al., \bibinfo{year}{2020}.
\newblock \bibinfo{title}{An image is worth 16x16 words: Transformers for image recognition at scale}.
\newblock \bibinfo{journal}{arXiv preprint arXiv:2010.11929} .
\bibitem[{Gao et~al.(2023)Gao, Zhang, Ma, Li and Zhang}]{gao2023correlation}
\bibinfo{author}{Gao, S.}, \bibinfo{author}{Zhang, Z.}, \bibinfo{author}{Ma, J.}, \bibinfo{author}{Li, Z.}, \bibinfo{author}{Zhang, S.}, \bibinfo{year}{2023}.
\newblock \bibinfo{title}{Correlation-aware mutual learning for semi-supervised medical image segmentation}, in: \bibinfo{booktitle}{International Conference on Medical Image Computing and Computer-Assisted Intervention}, \bibinfo{organization}{Springer}. pp. \bibinfo{pages}{98--108}.
\bibitem[{Guan et~al.(2019)Guan, Khan, Sikdar and Chitnis}]{guan2019fully}
\bibinfo{author}{Guan, S.}, \bibinfo{author}{Khan, A.A.}, \bibinfo{author}{Sikdar, S.}, \bibinfo{author}{Chitnis, P.V.}, \bibinfo{year}{2019}.
\newblock \bibinfo{title}{Fully dense unet for 2-d sparse photoacoustic tomography artifact removal}.
\newblock \bibinfo{journal}{IEEE journal of biomedical and health informatics} \bibinfo{volume}{24}, \bibinfo{pages}{568--576}.
\bibitem[{Hang et~al.(2020)Hang, Feng, Liang, Yu, Wang, Choi and Qin}]{hang2020local}
\bibinfo{author}{Hang, W.}, \bibinfo{author}{Feng, W.}, \bibinfo{author}{Liang, S.}, \bibinfo{author}{Yu, L.}, \bibinfo{author}{Wang, Q.}, \bibinfo{author}{Choi, K.S.}, \bibinfo{author}{Qin, J.}, \bibinfo{year}{2020}.
\newblock \bibinfo{title}{Local and global structure-aware entropy regularized mean teacher model for 3d left atrium segmentation}, in: \bibinfo{booktitle}{Medical Image Computing and Computer Assisted Intervention--MICCAI 2020: 23rd International Conference, Lima, Peru, October 4--8, 2020, Proceedings, Part I 23}, \bibinfo{organization}{Springer}. pp. \bibinfo{pages}{562--571}.
\bibitem[{He et~al.(2023)He, Wang, Li, Du, Xia and Fu}]{10093768}
\bibinfo{author}{He, A.}, \bibinfo{author}{Wang, K.}, \bibinfo{author}{Li, T.}, \bibinfo{author}{Du, C.}, \bibinfo{author}{Xia, S.}, \bibinfo{author}{Fu, H.}, \bibinfo{year}{2023}.
\newblock \bibinfo{title}{H2former: An efficient hierarchical hybrid transformer for medical image segmentation}.
\newblock \bibinfo{journal}{IEEE Transactions on Medical Imaging} \bibinfo{volume}{42}, \bibinfo{pages}{2763--2775}.
\newblock \DOIprefix\doi{10.1109/TMI.2023.3264513}.
\bibitem[{He et~al.(2020)He, Fan, Wu, Xie and Girshick}]{he2020momentum}
\bibinfo{author}{He, K.}, \bibinfo{author}{Fan, H.}, \bibinfo{author}{Wu, Y.}, \bibinfo{author}{Xie, S.}, \bibinfo{author}{Girshick, R.}, \bibinfo{year}{2020}.
\newblock \bibinfo{title}{Momentum contrast for unsupervised visual representation learning}, in: \bibinfo{booktitle}{Proceedings of the IEEE/CVF conference on computer vision and pattern recognition}, pp. \bibinfo{pages}{9729--9738}.
\bibitem[{Ho et~al.(2020)Ho, Jain and Abbeel}]{ho2020denoising}
\bibinfo{author}{Ho, J.}, \bibinfo{author}{Jain, A.}, \bibinfo{author}{Abbeel, P.}, \bibinfo{year}{2020}.
\newblock \bibinfo{title}{Denoising diffusion probabilistic models}.
\newblock \bibinfo{journal}{Advances in neural information processing systems} \bibinfo{volume}{33}, \bibinfo{pages}{6840--6851}.
\bibitem[{H{\"o}rst et~al.(2024)H{\"o}rst, Rempe, Heine, Seibold, Keyl, Baldini, Ugurel, Siveke, Gr{\"u}nwald, Egger et~al.}]{horst2024cellvit}
\bibinfo{author}{H{\"o}rst, F.}, \bibinfo{author}{Rempe, M.}, \bibinfo{author}{Heine, L.}, \bibinfo{author}{Seibold, C.}, \bibinfo{author}{Keyl, J.}, \bibinfo{author}{Baldini, G.}, \bibinfo{author}{Ugurel, S.}, \bibinfo{author}{Siveke, J.}, \bibinfo{author}{Gr{\"u}nwald, B.}, \bibinfo{author}{Egger, J.}, et~al., \bibinfo{year}{2024}.
\newblock \bibinfo{title}{Cellvit: Vision transformers for precise cell segmentation and classification}.
\newblock \bibinfo{journal}{Medical Image Analysis} , \bibinfo{pages}{103143}.
\bibitem[{Lei et~al.(2023)Lei, Liu, Wan, Li, Xia and Nandi}]{lei2023shape}
\bibinfo{author}{Lei, T.}, \bibinfo{author}{Liu, H.}, \bibinfo{author}{Wan, Y.}, \bibinfo{author}{Li, C.}, \bibinfo{author}{Xia, Y.}, \bibinfo{author}{Nandi, A.K.}, \bibinfo{year}{2023}.
\newblock \bibinfo{title}{Shape-guided dual consistency semi-supervised learning framework for 3d medical image segmentation}.
\newblock \bibinfo{journal}{IEEE Transactions on Radiation and Plasma Medical Sciences} .
\bibitem[{Lei et~al.(2022)Lei, Zhang, Du, Wang, Wan and Nandi}]{lei2022semi}
\bibinfo{author}{Lei, T.}, \bibinfo{author}{Zhang, D.}, \bibinfo{author}{Du, X.}, \bibinfo{author}{Wang, X.}, \bibinfo{author}{Wan, Y.}, \bibinfo{author}{Nandi, A.K.}, \bibinfo{year}{2022}.
\newblock \bibinfo{title}{Semi-supervised medical image segmentation using adversarial consistency learning and dynamic convolution network}.
\newblock \bibinfo{journal}{IEEE Transactions on Medical Imaging} .
\bibitem[{Li et~al.(2024)Li, Lian, Luo, Wang and Li}]{li2024contour}
\bibinfo{author}{Li, L.}, \bibinfo{author}{Lian, S.}, \bibinfo{author}{Luo, Z.}, \bibinfo{author}{Wang, B.}, \bibinfo{author}{Li, S.}, \bibinfo{year}{2024}.
\newblock \bibinfo{title}{Contour-aware consistency for semi-supervised medical image segmentation}.
\newblock \bibinfo{journal}{Biomedical Signal Processing and Control} \bibinfo{volume}{89}, \bibinfo{pages}{105694}.
\bibitem[{Li et~al.(2020a)Li, Zhang and He}]{li2020shape}
\bibinfo{author}{Li, S.}, \bibinfo{author}{Zhang, C.}, \bibinfo{author}{He, X.}, \bibinfo{year}{2020}a.
\newblock \bibinfo{title}{Shape-aware semi-supervised 3d semantic segmentation for medical images}, in: \bibinfo{booktitle}{Medical Image Computing and Computer Assisted Intervention--MICCAI 2020: 23rd International Conference, Lima, Peru, October 4--8, 2020, Proceedings, Part I 23}, \bibinfo{organization}{Springer}. pp. \bibinfo{pages}{552--561}.
\bibitem[{Li et~al.(2023)Li, Ding, Zhang, Yuan, Pang, Cheng, Chen, Liu and Loy}]{li2023transformer}
\bibinfo{author}{Li, X.}, \bibinfo{author}{Ding, H.}, \bibinfo{author}{Zhang, W.}, \bibinfo{author}{Yuan, H.}, \bibinfo{author}{Pang, J.}, \bibinfo{author}{Cheng, G.}, \bibinfo{author}{Chen, K.}, \bibinfo{author}{Liu, Z.}, \bibinfo{author}{Loy, C.C.}, \bibinfo{year}{2023}.
\newblock \bibinfo{title}{Transformer-based visual segmentation: A survey}.
\newblock \bibinfo{journal}{arXiv preprint arXiv:2304.09854} .
\bibitem[{Li et~al.(2020b)Li, Yu, Chen, Fu, Xing and Heng}]{li2020transformation}
\bibinfo{author}{Li, X.}, \bibinfo{author}{Yu, L.}, \bibinfo{author}{Chen, H.}, \bibinfo{author}{Fu, C.W.}, \bibinfo{author}{Xing, L.}, \bibinfo{author}{Heng, P.A.}, \bibinfo{year}{2020}b.
\newblock \bibinfo{title}{Transformation-consistent self-ensembling model for semisupervised medical image segmentation}.
\newblock \bibinfo{journal}{IEEE Transactions on Neural Networks and Learning Systems} \bibinfo{volume}{32}, \bibinfo{pages}{523--534}.
\bibitem[{Liu et~al.(2022a)Liu, Bao, Xie, Xiong, Sonke and Gavves}]{liu2022dynamic}
\bibinfo{author}{Liu, J.}, \bibinfo{author}{Bao, Y.}, \bibinfo{author}{Xie, G.S.}, \bibinfo{author}{Xiong, H.}, \bibinfo{author}{Sonke, J.J.}, \bibinfo{author}{Gavves, E.}, \bibinfo{year}{2022}a.
\newblock \bibinfo{title}{Dynamic prototype convolution network for few-shot semantic segmentation}, in: \bibinfo{booktitle}{Proceedings of the IEEE/CVF Conference on Computer Vision and Pattern Recognition}, pp. \bibinfo{pages}{11553--11562}.
\bibitem[{Liu et~al.(2021)Liu, Zhi, Johns and Davison}]{liu2021bootstrapping}
\bibinfo{author}{Liu, S.}, \bibinfo{author}{Zhi, S.}, \bibinfo{author}{Johns, E.}, \bibinfo{author}{Davison, A.J.}, \bibinfo{year}{2021}.
\newblock \bibinfo{title}{Bootstrapping semantic segmentation with regional contrast}.
\newblock \bibinfo{journal}{arXiv preprint arXiv:2104.04465} .
\bibitem[{Liu et~al.(2022b)Liu, Tian, Chen, Liu, Belagiannis and Carneiro}]{liu2022perturbed}
\bibinfo{author}{Liu, Y.}, \bibinfo{author}{Tian, Y.}, \bibinfo{author}{Chen, Y.}, \bibinfo{author}{Liu, F.}, \bibinfo{author}{Belagiannis, V.}, \bibinfo{author}{Carneiro, G.}, \bibinfo{year}{2022}b.
\newblock \bibinfo{title}{Perturbed and strict mean teachers for semi-supervised semantic segmentation}, in: \bibinfo{booktitle}{Proceedings of the IEEE/CVF Conference on Computer Vision and Pattern Recognition}, pp. \bibinfo{pages}{4258--4267}.
\bibitem[{Liu et~al.(2022c)Liu, Tian, Wang, Chen, Liu, Belagiannis and Carneiro}]{liu2022translation}
\bibinfo{author}{Liu, Y.}, \bibinfo{author}{Tian, Y.}, \bibinfo{author}{Wang, C.}, \bibinfo{author}{Chen, Y.}, \bibinfo{author}{Liu, F.}, \bibinfo{author}{Belagiannis, V.}, \bibinfo{author}{Carneiro, G.}, \bibinfo{year}{2022}c.
\newblock \bibinfo{title}{Translation consistent semi-supervised segmentation for 3d medical images}.
\newblock \bibinfo{journal}{arXiv preprint arXiv:2203.14523} .
\bibitem[{Lu et~al.(2024)Lu, Yan, Chen, Cheng, Zhang and Yang}]{LU2024107991}
\bibinfo{author}{Lu, S.}, \bibinfo{author}{Yan, Z.}, \bibinfo{author}{Chen, W.}, \bibinfo{author}{Cheng, T.}, \bibinfo{author}{Zhang, Z.}, \bibinfo{author}{Yang, G.}, \bibinfo{year}{2024}.
\newblock \bibinfo{title}{Dual consistency regularization with subjective logic for semi-supervised medical image segmentation}.
\newblock \bibinfo{journal}{Computers in Biology and Medicine} \bibinfo{volume}{170}, \bibinfo{pages}{107991}.
\newblock \URLprefix \url{https://www.sciencedirect.com/science/article/pii/S0010482524000751}, \DOIprefix\doi{https://doi.org/10.1016/j.compbiomed.2024.107991}.
\bibitem[{Luo et~al.(2021a)Luo, Chen, Song and Wang}]{luo2021semi}
\bibinfo{author}{Luo, X.}, \bibinfo{author}{Chen, J.}, \bibinfo{author}{Song, T.}, \bibinfo{author}{Wang, G.}, \bibinfo{year}{2021}a.
\newblock \bibinfo{title}{Semi-supervised medical image segmentation through dual-task consistency}, in: \bibinfo{booktitle}{Proceedings of the AAAI conference on artificial intelligence}, pp. \bibinfo{pages}{8801--8809}.
\bibitem[{Luo et~al.(2021b)Luo, Liao, Chen, Song, Chen, Zhang, Chen, Wang and Zhang}]{luo2021efficient}
\bibinfo{author}{Luo, X.}, \bibinfo{author}{Liao, W.}, \bibinfo{author}{Chen, J.}, \bibinfo{author}{Song, T.}, \bibinfo{author}{Chen, Y.}, \bibinfo{author}{Zhang, S.}, \bibinfo{author}{Chen, N.}, \bibinfo{author}{Wang, G.}, \bibinfo{author}{Zhang, S.}, \bibinfo{year}{2021}b.
\newblock \bibinfo{title}{Efficient semi-supervised gross target volume of nasopharyngeal carcinoma segmentation via uncertainty rectified pyramid consistency}, in: \bibinfo{booktitle}{Medical Image Computing and Computer Assisted Intervention--MICCAI 2021: 24th International Conference, Strasbourg, France, September 27--October 1, 2021, Proceedings, Part II 24}, \bibinfo{organization}{Springer}. pp. \bibinfo{pages}{318--329}.
\bibitem[{Menze et~al.(2014)Menze, Jakab, Bauer, Kalpathy-Cramer, Farahani, Kirby, Burren, Porz, Slotboom, Wiest et~al.}]{menze2014multimodal}
\bibinfo{author}{Menze, B.H.}, \bibinfo{author}{Jakab, A.}, \bibinfo{author}{Bauer, S.}, \bibinfo{author}{Kalpathy-Cramer, J.}, \bibinfo{author}{Farahani, K.}, \bibinfo{author}{Kirby, J.}, \bibinfo{author}{Burren, Y.}, \bibinfo{author}{Porz, N.}, \bibinfo{author}{Slotboom, J.}, \bibinfo{author}{Wiest, R.}, et~al., \bibinfo{year}{2014}.
\newblock \bibinfo{title}{The multimodal brain tumor image segmentation benchmark (brats)}.
\newblock \bibinfo{journal}{IEEE transactions on medical imaging} \bibinfo{volume}{34}, \bibinfo{pages}{1993--2024}.
\bibitem[{Miao et~al.(2023)Miao, Chen, Liu, Wei and Heng}]{miao2023caussl}
\bibinfo{author}{Miao, J.}, \bibinfo{author}{Chen, C.}, \bibinfo{author}{Liu, F.}, \bibinfo{author}{Wei, H.}, \bibinfo{author}{Heng, P.A.}, \bibinfo{year}{2023}.
\newblock \bibinfo{title}{Caussl: Causality-inspired semi-supervised learning for medical image segmentation}, in: \bibinfo{booktitle}{Proceedings of the IEEE/CVF International Conference on Computer Vision}, pp. \bibinfo{pages}{21426--21437}.
\bibitem[{Milletari et~al.(2016)Milletari, Navab and Ahmadi}]{milletari2016v}
\bibinfo{author}{Milletari, F.}, \bibinfo{author}{Navab, N.}, \bibinfo{author}{Ahmadi, S.A.}, \bibinfo{year}{2016}.
\newblock \bibinfo{title}{V-net: Fully convolutional neural networks for volumetric medical image segmentation}, in: \bibinfo{booktitle}{2016 fourth international conference on 3D vision (3DV)}, \bibinfo{organization}{Ieee}. pp. \bibinfo{pages}{565--571}.
\bibitem[{Minaee et~al.(2021)Minaee, Boykov, Porikli, Plaza, Kehtarnavaz and Terzopoulos}]{minaee2021image}
\bibinfo{author}{Minaee, S.}, \bibinfo{author}{Boykov, Y.}, \bibinfo{author}{Porikli, F.}, \bibinfo{author}{Plaza, A.}, \bibinfo{author}{Kehtarnavaz, N.}, \bibinfo{author}{Terzopoulos, D.}, \bibinfo{year}{2021}.
\newblock \bibinfo{title}{Image segmentation using deep learning: A survey}.
\newblock \bibinfo{journal}{IEEE transactions on pattern analysis and machine intelligence} \bibinfo{volume}{44}, \bibinfo{pages}{3523--3542}.
\bibitem[{Ronneberger et~al.(2015)Ronneberger, Fischer and Brox}]{ronneberger2015u}
\bibinfo{author}{Ronneberger, O.}, \bibinfo{author}{Fischer, P.}, \bibinfo{author}{Brox, T.}, \bibinfo{year}{2015}.
\newblock \bibinfo{title}{U-net: Convolutional networks for biomedical image segmentation}, in: \bibinfo{booktitle}{Medical Image Computing and Computer-Assisted Intervention--MICCAI 2015: 18th International Conference, Munich, Germany, October 5-9, 2015, Proceedings, Part III 18}, \bibinfo{organization}{Springer}. pp. \bibinfo{pages}{234--241}.
\bibitem[{Shaharabany et~al.(2023)Shaharabany, Dahan, Giryes and Wolf}]{shaharabany2023autosam}
\bibinfo{author}{Shaharabany, T.}, \bibinfo{author}{Dahan, A.}, \bibinfo{author}{Giryes, R.}, \bibinfo{author}{Wolf, L.}, \bibinfo{year}{2023}.
\newblock \bibinfo{title}{Autosam: Adapting sam to medical images by overloading the prompt encoder}.
\newblock \bibinfo{journal}{arXiv preprint arXiv:2306.06370} .
\bibitem[{Su et~al.(2024)Su, Luo, Lian, Lin and Li}]{su2024mutual}
\bibinfo{author}{Su, J.}, \bibinfo{author}{Luo, Z.}, \bibinfo{author}{Lian, S.}, \bibinfo{author}{Lin, D.}, \bibinfo{author}{Li, S.}, \bibinfo{year}{2024}.
\newblock \bibinfo{title}{Mutual learning with reliable pseudo label for semi-supervised medical image segmentation}.
\newblock \bibinfo{journal}{Medical Image Analysis} , \bibinfo{pages}{103111}.
\bibitem[{Wang et~al.(2021)Wang, Zhan, Zu, Wu, Zhou, Zhou and Wang}]{wang2021tripled}
\bibinfo{author}{Wang, K.}, \bibinfo{author}{Zhan, B.}, \bibinfo{author}{Zu, C.}, \bibinfo{author}{Wu, X.}, \bibinfo{author}{Zhou, J.}, \bibinfo{author}{Zhou, L.}, \bibinfo{author}{Wang, Y.}, \bibinfo{year}{2021}.
\newblock \bibinfo{title}{Tripled-uncertainty guided mean teacher model for semi-supervised medical image segmentation}, in: \bibinfo{booktitle}{Medical Image Computing and Computer Assisted Intervention--MICCAI 2021: 24th International Conference, Strasbourg, France, September 27--October 1, 2021, Proceedings, Part II 24}, \bibinfo{organization}{Springer}. pp. \bibinfo{pages}{450--460}.
\bibitem[{Wang et~al.(2022)Wang, Lei, Cui, Zhang, Meng and Nandi}]{wang2022medical}
\bibinfo{author}{Wang, R.}, \bibinfo{author}{Lei, T.}, \bibinfo{author}{Cui, R.}, \bibinfo{author}{Zhang, B.}, \bibinfo{author}{Meng, H.}, \bibinfo{author}{Nandi, A.K.}, \bibinfo{year}{2022}.
\newblock \bibinfo{title}{Medical image segmentation using deep learning: A survey}.
\newblock \bibinfo{journal}{IET Image Processing} \bibinfo{volume}{16}, \bibinfo{pages}{1243--1267}.
\bibitem[{Wang et~al.(2023)Wang, Xiao, Bi, Li and Gao}]{wang2023mcf}
\bibinfo{author}{Wang, Y.}, \bibinfo{author}{Xiao, B.}, \bibinfo{author}{Bi, X.}, \bibinfo{author}{Li, W.}, \bibinfo{author}{Gao, X.}, \bibinfo{year}{2023}.
\newblock \bibinfo{title}{Mcf: Mutual correction framework for semi-supervised medical image segmentation}, in: \bibinfo{booktitle}{Proceedings of the IEEE/CVF Conference on Computer Vision and Pattern Recognition}, pp. \bibinfo{pages}{15651--15660}.
\bibitem[{Wang et~al.(2020)Wang, Zhang, Tian, Zhong, Shi, Zhang and He}]{wang2020double}
\bibinfo{author}{Wang, Y.}, \bibinfo{author}{Zhang, Y.}, \bibinfo{author}{Tian, J.}, \bibinfo{author}{Zhong, C.}, \bibinfo{author}{Shi, Z.}, \bibinfo{author}{Zhang, Y.}, \bibinfo{author}{He, Z.}, \bibinfo{year}{2020}.
\newblock \bibinfo{title}{Double-uncertainty weighted method for semi-supervised learning}, in: \bibinfo{booktitle}{Medical Image Computing and Computer Assisted Intervention--MICCAI 2020: 23rd International Conference, Lima, Peru, October 4--8, 2020, Proceedings, Part I 23}, \bibinfo{organization}{Springer}. pp. \bibinfo{pages}{542--551}.
\bibitem[{Wu et~al.(2024)Wu, Ji, Fu, Xu, Jin and Xu}]{wu2024medsegdiff}
\bibinfo{author}{Wu, J.}, \bibinfo{author}{Ji, W.}, \bibinfo{author}{Fu, H.}, \bibinfo{author}{Xu, M.}, \bibinfo{author}{Jin, Y.}, \bibinfo{author}{Xu, Y.}, \bibinfo{year}{2024}.
\newblock \bibinfo{title}{Medsegdiff-v2: Diffusion-based medical image segmentation with transformer}, in: \bibinfo{booktitle}{Proceedings of the AAAI Conference on Artificial Intelligence}, pp. \bibinfo{pages}{6030--6038}.
\bibitem[{Wu et~al.(2022)Wu, Ge, Zhang, Xu, Zhang, Xia and Cai}]{wu2022mutual}
\bibinfo{author}{Wu, Y.}, \bibinfo{author}{Ge, Z.}, \bibinfo{author}{Zhang, D.}, \bibinfo{author}{Xu, M.}, \bibinfo{author}{Zhang, L.}, \bibinfo{author}{Xia, Y.}, \bibinfo{author}{Cai, J.}, \bibinfo{year}{2022}.
\newblock \bibinfo{title}{Mutual consistency learning for semi-supervised medical image segmentation}.
\newblock \bibinfo{journal}{Medical Image Analysis} \bibinfo{volume}{81}, \bibinfo{pages}{102530}.
\bibitem[{Wu et~al.(2021)Wu, Xu, Ge, Cai and Zhang}]{wu2021semi}
\bibinfo{author}{Wu, Y.}, \bibinfo{author}{Xu, M.}, \bibinfo{author}{Ge, Z.}, \bibinfo{author}{Cai, J.}, \bibinfo{author}{Zhang, L.}, \bibinfo{year}{2021}.
\newblock \bibinfo{title}{Semi-supervised left atrium segmentation with mutual consistency training}, in: \bibinfo{booktitle}{Medical Image Computing and Computer Assisted Intervention--MICCAI 2021: 24th International Conference, Strasbourg, France, September 27--October 1, 2021, Proceedings, Part II 24}, \bibinfo{organization}{Springer}. pp. \bibinfo{pages}{297--306}.
\bibitem[{Xia et~al.(2020)Xia, Yang, Yu, Liu, Cai, Yu, Zhu, Xu, Yuille and Roth}]{xia2020uncertainty}
\bibinfo{author}{Xia, Y.}, \bibinfo{author}{Yang, D.}, \bibinfo{author}{Yu, Z.}, \bibinfo{author}{Liu, F.}, \bibinfo{author}{Cai, J.}, \bibinfo{author}{Yu, L.}, \bibinfo{author}{Zhu, Z.}, \bibinfo{author}{Xu, D.}, \bibinfo{author}{Yuille, A.}, \bibinfo{author}{Roth, H.}, \bibinfo{year}{2020}.
\newblock \bibinfo{title}{Uncertainty-aware multi-view co-training for semi-supervised medical image segmentation and domain adaptation}.
\newblock \bibinfo{journal}{Medical image analysis} \bibinfo{volume}{65}, \bibinfo{pages}{101766}.
\bibitem[{Xiao et~al.(2018)Xiao, Lian, Luo and Li}]{xiao2018weighted}
\bibinfo{author}{Xiao, X.}, \bibinfo{author}{Lian, S.}, \bibinfo{author}{Luo, Z.}, \bibinfo{author}{Li, S.}, \bibinfo{year}{2018}.
\newblock \bibinfo{title}{Weighted res-unet for high-quality retina vessel segmentation}, in: \bibinfo{booktitle}{2018 9th international conference on information technology in medicine and education (ITME)}, \bibinfo{organization}{IEEE}. pp. \bibinfo{pages}{327--331}.
\bibitem[{Xiong et~al.(2021)Xiong, Xia, Hu, Huang, Bian, Zheng, Vesal, Ravikumar, Maier, Yang et~al.}]{xiong2021global}
\bibinfo{author}{Xiong, Z.}, \bibinfo{author}{Xia, Q.}, \bibinfo{author}{Hu, Z.}, \bibinfo{author}{Huang, N.}, \bibinfo{author}{Bian, C.}, \bibinfo{author}{Zheng, Y.}, \bibinfo{author}{Vesal, S.}, \bibinfo{author}{Ravikumar, N.}, \bibinfo{author}{Maier, A.}, \bibinfo{author}{Yang, X.}, et~al., \bibinfo{year}{2021}.
\newblock \bibinfo{title}{A global benchmark of algorithms for segmenting the left atrium from late gadolinium-enhanced cardiac magnetic resonance imaging}.
\newblock \bibinfo{journal}{Medical image analysis} \bibinfo{volume}{67}, \bibinfo{pages}{101832}.
\bibitem[{Xu et~al.(2022)Xu, Zhou, Jin, Blumberg, Wilson, deGroot, Alexander, Oxtoby and Jacob}]{xu2022learning}
\bibinfo{author}{Xu, M.C.}, \bibinfo{author}{Zhou, Y.K.}, \bibinfo{author}{Jin, C.}, \bibinfo{author}{Blumberg, S.B.}, \bibinfo{author}{Wilson, F.J.}, \bibinfo{author}{deGroot, M.}, \bibinfo{author}{Alexander, D.C.}, \bibinfo{author}{Oxtoby, N.P.}, \bibinfo{author}{Jacob, J.}, \bibinfo{year}{2022}.
\newblock \bibinfo{title}{Learning morphological feature perturbations for calibrated semi-supervised segmentation}, in: \bibinfo{booktitle}{International Conference on Medical Imaging with Deep Learning}, \bibinfo{organization}{PMLR}. pp. \bibinfo{pages}{1413--1429}.
\bibitem[{Yu et~al.(2019)Yu, Wang, Li, Fu and Heng}]{yu2019uncertainty}
\bibinfo{author}{Yu, L.}, \bibinfo{author}{Wang, S.}, \bibinfo{author}{Li, X.}, \bibinfo{author}{Fu, C.W.}, \bibinfo{author}{Heng, P.A.}, \bibinfo{year}{2019}.
\newblock \bibinfo{title}{Uncertainty-aware self-ensembling model for semi-supervised 3d left atrium segmentation}, in: \bibinfo{booktitle}{Medical Image Computing and Computer Assisted Intervention--MICCAI 2019: 22nd International Conference, Shenzhen, China, October 13--17, 2019, Proceedings, Part II 22}, \bibinfo{organization}{Springer}. pp. \bibinfo{pages}{605--613}.
\bibitem[{Yun et~al.(2019)Yun, Han, Oh, Chun, Choe and Yoo}]{yun2019cutmix}
\bibinfo{author}{Yun, S.}, \bibinfo{author}{Han, D.}, \bibinfo{author}{Oh, S.J.}, \bibinfo{author}{Chun, S.}, \bibinfo{author}{Choe, J.}, \bibinfo{author}{Yoo, Y.}, \bibinfo{year}{2019}.
\newblock \bibinfo{title}{Cutmix: Regularization strategy to train strong classifiers with localizable features}, in: \bibinfo{booktitle}{Proceedings of the IEEE/CVF international conference on computer vision}, pp. \bibinfo{pages}{6023--6032}.
\bibitem[{Zhang et~al.(2022)Zhang, Tian, Gao, Wang, Feng, Bai and Jiao}]{ZHANG2022369}
\bibinfo{author}{Zhang, Z.}, \bibinfo{author}{Tian, C.}, \bibinfo{author}{Gao, X.}, \bibinfo{author}{Wang, C.}, \bibinfo{author}{Feng, X.}, \bibinfo{author}{Bai, H.X.}, \bibinfo{author}{Jiao, Z.}, \bibinfo{year}{2022}.
\newblock \bibinfo{title}{Dynamic prototypical feature representation learning framework for semi-supervised skin lesion segmentation}.
\newblock \bibinfo{journal}{Neurocomputing} \bibinfo{volume}{507}, \bibinfo{pages}{369--382}.
\newblock \URLprefix \url{https://www.sciencedirect.com/science/article/pii/S092523122201013X}, \DOIprefix\doi{https://doi.org/10.1016/j.neucom.2022.08.039}.
\bibitem[{Zhao et~al.(2024)Zhao, Qi, Wang, Wang, Wu, Mao and Zhang}]{10273222}
\bibinfo{author}{Zhao, X.}, \bibinfo{author}{Qi, Z.}, \bibinfo{author}{Wang, S.}, \bibinfo{author}{Wang, Q.}, \bibinfo{author}{Wu, X.}, \bibinfo{author}{Mao, Y.}, \bibinfo{author}{Zhang, L.}, \bibinfo{year}{2024}.
\newblock \bibinfo{title}{Rcps: Rectified contrastive pseudo supervision for semi-supervised medical image segmentation}.
\newblock \bibinfo{journal}{IEEE Journal of Biomedical and Health Informatics} \bibinfo{volume}{28}, \bibinfo{pages}{251--261}.
\newblock \DOIprefix\doi{10.1109/JBHI.2023.3322590}.
\bibitem[{Zhao et~al.(2023)Zhao, Wang, Wang, Yuan and Zhou}]{zhao2023alternate}
\bibinfo{author}{Zhao, Z.}, \bibinfo{author}{Wang, Z.}, \bibinfo{author}{Wang, L.}, \bibinfo{author}{Yuan, Y.}, \bibinfo{author}{Zhou, L.}, \bibinfo{year}{2023}.
\newblock \bibinfo{title}{Alternate diverse teaching for semi-supervised medical image segmentation}.
\newblock \bibinfo{journal}{arXiv preprint arXiv:2311.17325} .
\bibitem[{Zheng et~al.(2022)Zheng, Xu and Wei}]{zheng2022double}
\bibinfo{author}{Zheng, K.}, \bibinfo{author}{Xu, J.}, \bibinfo{author}{Wei, J.}, \bibinfo{year}{2022}.
\newblock \bibinfo{title}{Double noise mean teacher self-ensembling model for semi-supervised tumor segmentation}, in: \bibinfo{booktitle}{ICASSP 2022-2022 IEEE International Conference on Acoustics, Speech and Signal Processing (ICASSP)}, \bibinfo{organization}{IEEE}. pp. \bibinfo{pages}{1446--1450}.
\bibitem[{Zhou et~al.(2018)Zhou, Rahman~Siddiquee, Tajbakhsh and Liang}]{zhou2018unet++}
\bibinfo{author}{Zhou, Z.}, \bibinfo{author}{Rahman~Siddiquee, M.M.}, \bibinfo{author}{Tajbakhsh, N.}, \bibinfo{author}{Liang, J.}, \bibinfo{year}{2018}.
\newblock \bibinfo{title}{Unet++: A nested u-net architecture for medical image segmentation}, in: \bibinfo{booktitle}{Deep Learning in Medical Image Analysis and Multimodal Learning for Clinical Decision Support: 4th International Workshop, DLMIA 2018, and 8th International Workshop, ML-CDS 2018, Held in Conjunction with MICCAI 2018, Granada, Spain, September 20, 2018, Proceedings 4}, \bibinfo{organization}{Springer}. pp. \bibinfo{pages}{3--11}.

\end{thebibliography}

\end{document}